\documentclass[journal,comsoc]{IEEEtran}

\usepackage[T1]{fontenc}

\usepackage{ifpdf}
\usepackage{cite}
\usepackage{graphicx}
\graphicspath{{./Fig/}}
\usepackage{amsmath,amsfonts}
\usepackage{bm}
\usepackage{algorithm}
\usepackage{algorithmic}
\usepackage{array}
\usepackage[caption=false,font=normalsize,labelfont=sf,textfont=sf]{subfig}
\usepackage{url}
\begin{document}

\title{Design and Characterization of MRI-compatible Plastic Ultrasonic Motor}

\author{Zhanyue Zhao, Charles Bales, Gregory Fischer
\thanks{Zhanyue Zhao, Charles Bales, and Gregory Fischer are with the Department of Robotics Engineering, Worcester Polytechnic Institute, Worcester, MA 01605 USA (e-mail: zzhao4@wpi.edu, gfischer@wpi.edu).}
\thanks{This research is supported by National Institute of Health (NIH) under the National Cancer Institute (NCI) under Grant R01CA166379 and R01EB030539.}
}


\maketitle

\begin{abstract}

Precise surgical procedures may benefit from intra-operative image guidance using magnetic resonance imaging (MRI). However, the MRI’s strong magnetic fields, fast switching gradients, and constrained space pose the need for an MR-guided robotic system to assist the surgeon. Piezoelectric actuators can be used in an MRI environment by utilizing the inverse piezoelectric effect for different application purposes. Piezoelectric ultrasonic motor (USM) is one type of MRI-compatible actuator that can actuate these robots with fast response times, compactness, and simple configuration. Although the piezoelectric motors are mostly made of nonferromagnetic material, the generation of eddy currents due to the MRI’s gradient fields can lead to magnetic field distortions causing image artifacts. Motor vibrations due to interactions between the MRI’s magnetic fields and those generated by the eddy currents can further degrade image quality by causing image artifacts. In this work, a plastic piezoelectric ultrasonic (USM) motor with more degree of MRI compatibility was developed and induced with preliminary optimization. Multiple parameters, namely teeth number, notch size, edge bevel or straight, and surface finish level parameters were used versus the prepressure for the experiment, and the results suggested that using 48 teeth, thin teeth notch with 0.39mm, beveled edge and a surface finish using grit number of approximate 1000 sandpaper performed a better output both in rotary speed and torque. Under this combination, the highest speed reached up to 436.6665rpm when the prepressure was low, and the highest torque reached up to 0.0348Nm when the prepressure was approximately 500g.

\end{abstract}

\begin{IEEEkeywords}

MRI Compatible Actuator, Plastic Ultrasonic Motor, Piezoelectric Actuator

\end{IEEEkeywords}

\section{Introduction}

\IEEEPARstart{A}{ctuator} selection and design is the most critical section for MRI-compatible robotic systems, given actuator is the core component for robot actuation and usually decides the class of MRI compatibility of the robot. The actuator decides the size and structure of the robotic system which ensures the feasibility of fitting into the limited MRI bore space \cite{su2022state}. Another consideration is to minimize electromagnetic interference (EMI) and avoid image artifacts and drop in signal-to-noise ratio (SNR). Conventional actuators such as DC motors can not be used in an MRI environment due to the magnets, coils of wire, and ferrous enclosures that can pose dangerous projectiles near the strong magnetic field of an MRI machine. The piezoelectric motors are operated based on the geometric change of a piezoelectric material to generate motion under an electric field applied \cite{xiao2020mr}. Due to their high precision, fast response times, compactness, and simpler system setup, piezoelectric motors are favorable for MRI-compatible applications \cite{su2022state}, and it is the only commercially available actuators to be used in the MRI environment, which is widely used in our related study using MRI-compatible robotic system aiming various studies \cite{campwala2021predicting,gandomi2020modeling, szewczyk2022happens,tavakkolmoghaddam2021neuroplan,tavakkolmoghaddam2023passive, tavakkolmoghaddam2023design}.

A large group of research efforts has been dedicated to characterizing and improving its coexistence with MRI equipment. Park \textit{et al.} developed a spherical actuator \cite{takemura2001design} which they then extended into an actuator for a laparoscopic instrument \cite{park2004study}. Their actuator is a sphere that is frictionally connected to a piezoelectric stack. The sphere rotates in a controlled manner when the proper vibration modes are induced. Due to the magnetic pre-loading of the sphere onto the piezoelectric stack, its prototypes are incompatible with the MRI environment. However, they point out that using an elastic-based loading can overcome this problem. Elhawary \textit{et al.} demonstrated the use of a non-resonant PiezoLegs (Micromo, FL, United States) actuator in a 1 DOF freedom prototype in an MRI. Their system was able to achieve 36 um accuracy with a maximum SNR reduction of 27.6\% using a True Fisp gradient echo sequence (FOV: 230mm, TR/TE: 6.46/3.05ms, 256$\times$256, FA: 80$^\circ$, 6 slices, slice thickness and spacing: 5mm) \cite{elhawary2006feasibility}. Chapuis \textit{et al.} combine a USR30-E3N (Shinsei, Japan) with a clutch to build a haptic knob device that is not designed for use in an MRI \cite{chapuis2007haptic}. Mashimo \textit{et al.} also developed a spherical actuator \cite{mashimo2007mri}. Their design utilizes 3 stators similar to those used by regular resonant motors to enable the spinning of the sphere. Khanicheh \textit{et al.'s} work also collaborates the electrorheological fluid-based devices compatibility with operation in the MR environment \cite{khanicheh2008evaluation}. Piezoelectric Actuator Drive (PAD) is one of the piezoelectric-driven actuators.  Vogle \textit{et al.} built upon the PAD design presented \cite{kappel2008piezoelektrischer} to create a version compatible with use inside the MRI environment \cite{vogl2009new}. However, their steel construction causes significant image distortion requiring that the actuator be at least 20cm from the region of interest. Mangeot \textit{et al.} built a motor based on the PAD architecture and tested it in a Siemens 3T MRI machine, and the motor can be used very close to the imaging volume without significant loss of performance and without affecting the medical diagnostic image \cite{mangeot2014design}. Rotary-linear combined actuation is another group of piezoelectric actuators. Mashimo \textit{et al.} built a piezoelectric actuator having a single stator with rotary and linear motions. The stator was able to rotate and translate the shaft at the frequency of the R3 mode and at the common frequency of the T1 and T2 modes. The maximum rotational speed of the shaft was approximately 160 rpm, and the maximum linear speed was about 63mm/s at an applied voltage of 42$V_{rms}$ \cite{mashimo2009rotary}. Tuncdemir \textit{et al.} designed a multi-degree-of-freedom (MDoF) ultrasonic motor by using a single actuator that combined rotary and translational motions in a cylindrical-joint type configuration. The prototype of the motor (5mm diameter, 25mm total length) has 5 mm/s translational and 3 rad/s rotary speed under 4mN blocking force when the input signal is 20$V_{pp}$ square wave \cite{tuncdemir2011design}. Elbannan \textit{et al.} designed a 2 DOF inchworm actuator \cite{elbannan2012design} based on the concepts introduced in \cite{salisbury2006design}. Their constructed prototype \cite{el2015development} is compatible with use inside an MRI. The actuator is capable of 5.4mm/s linear motion and 10.5rpm rotary motion.

Despite their advantages, piezoelectric actuators have certain limitations. The process of converting electrical energy into mechanical motion restricts these actuators to the MRI-conditional class, unlike pneumatic and hydraulic actuators, which can be manufactured entirely from MRI-safe materials. Furthermore, commercial piezoelectric motors and drivers can cause up to an 80\% signal loss during synchronous robot motion. Even with the use of well-designed RF shielding sleeves and properly grounded control cables, significant signal-to-noise ratio (SNR) reduction persists \cite{su2022state}. Although piezoelectric motors are generally constructed from nonferromagnetic materials, the MRI’s gradient fields can induce eddy currents, leading to magnetic field distortions and image artifacts. Additionally, motor vibrations resulting from the interaction between the MRI’s magnetic fields and the eddy currents can further degrade image quality by introducing artifacts. Carvalho \textit{et al.} demonstrated that by replacing the non-core components, such as the metallic housing and shaft of a commercially available USR60 piezoelectric motor, with plastic materials and using an appropriate controller, the modified actuator could be used in MRI scanners with reduced noise and paramagnetic artifacts \cite{carvalho2020demonstration}. However, the stator, a key component, remained metallic (made of copper), meaning that full MRI compatibility was not achieved. In this work, we build upon our previous research \cite{carvalho2020demonstration, zhao2021preliminary,zhao2023preliminary,zhao2024study} by further demetallizing the motor, replacing the core components with plastic materials, and developing a fully plastic ultrasonic motor (USM) that offers a higher degree of MRI compatibility.

\section{Design and Configuration}

\subsection{Plastic USM Whole Motor Design}

\begin{figure}
    \centering
    \includegraphics[width=0.9\linewidth]{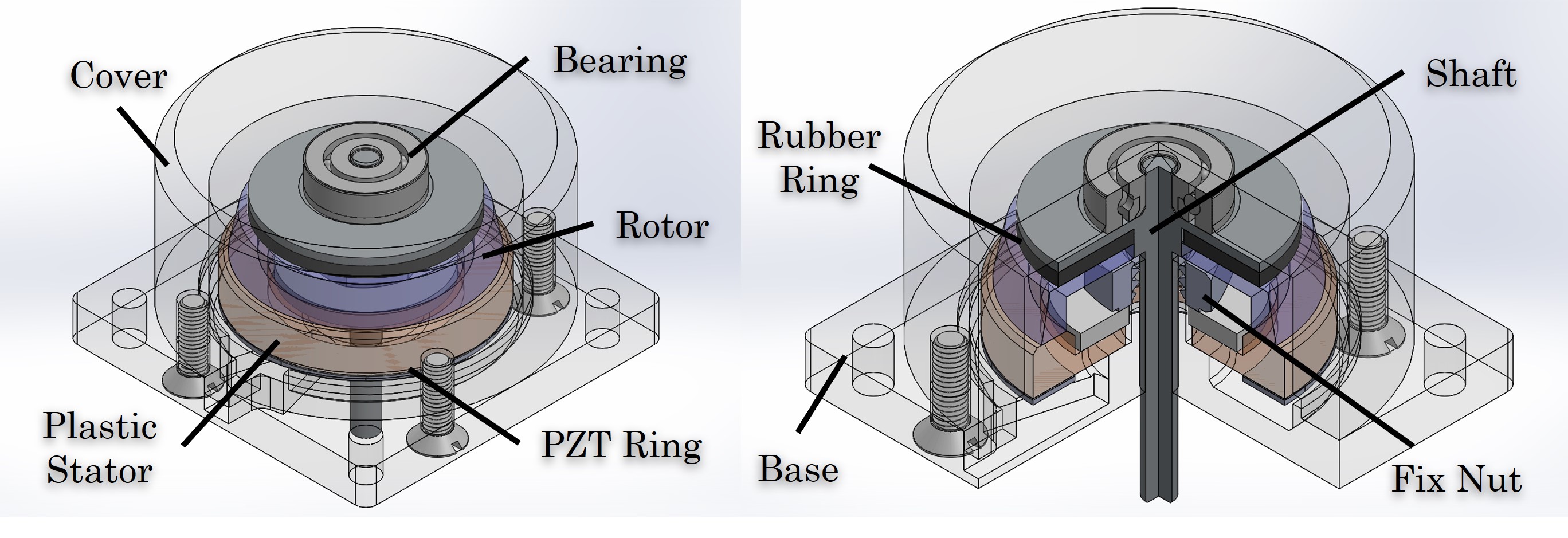}
    \caption{The CAD drawing of the MRI-compatible USM.}
    \label{fig:PlasticDesign}
\end{figure}

\begin{figure}
    \centering
    \includegraphics[width=1\linewidth]{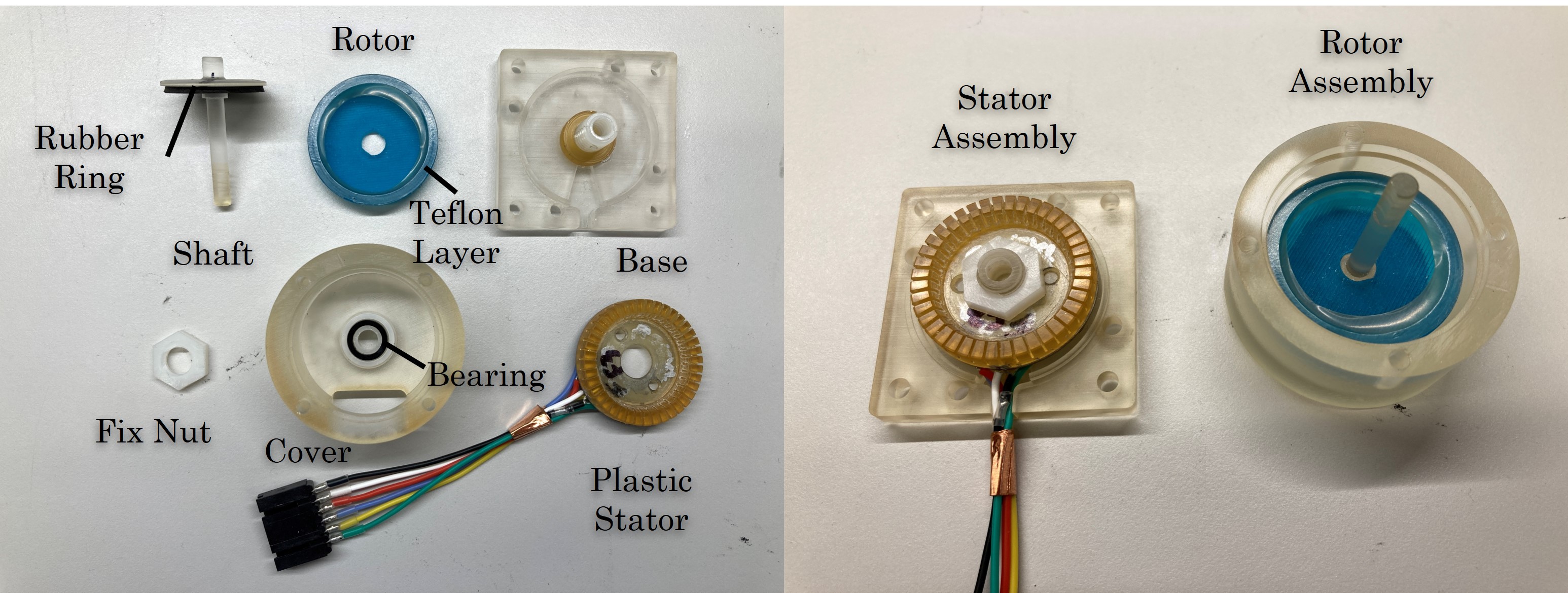}
    \caption{Components of MRI-compatible USM.}
    \label{fig:PMpart}
\end{figure}

\begin{figure}
    \centering
    \includegraphics[width=0.8\linewidth]{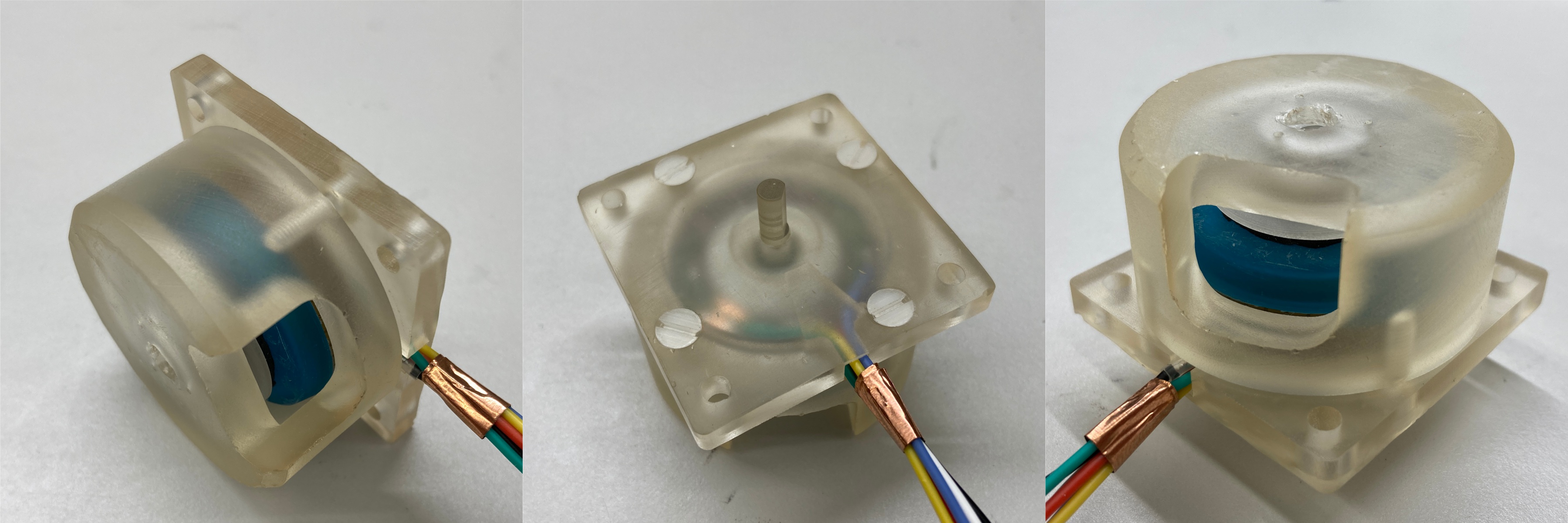}
    \caption{Enclosed and finished status of the motor. A hole at the side was for rotary motion observation.}
    \label{fig:PMassem}
\end{figure}

\begin{figure}
    \centering
    \includegraphics[width=0.8\linewidth]{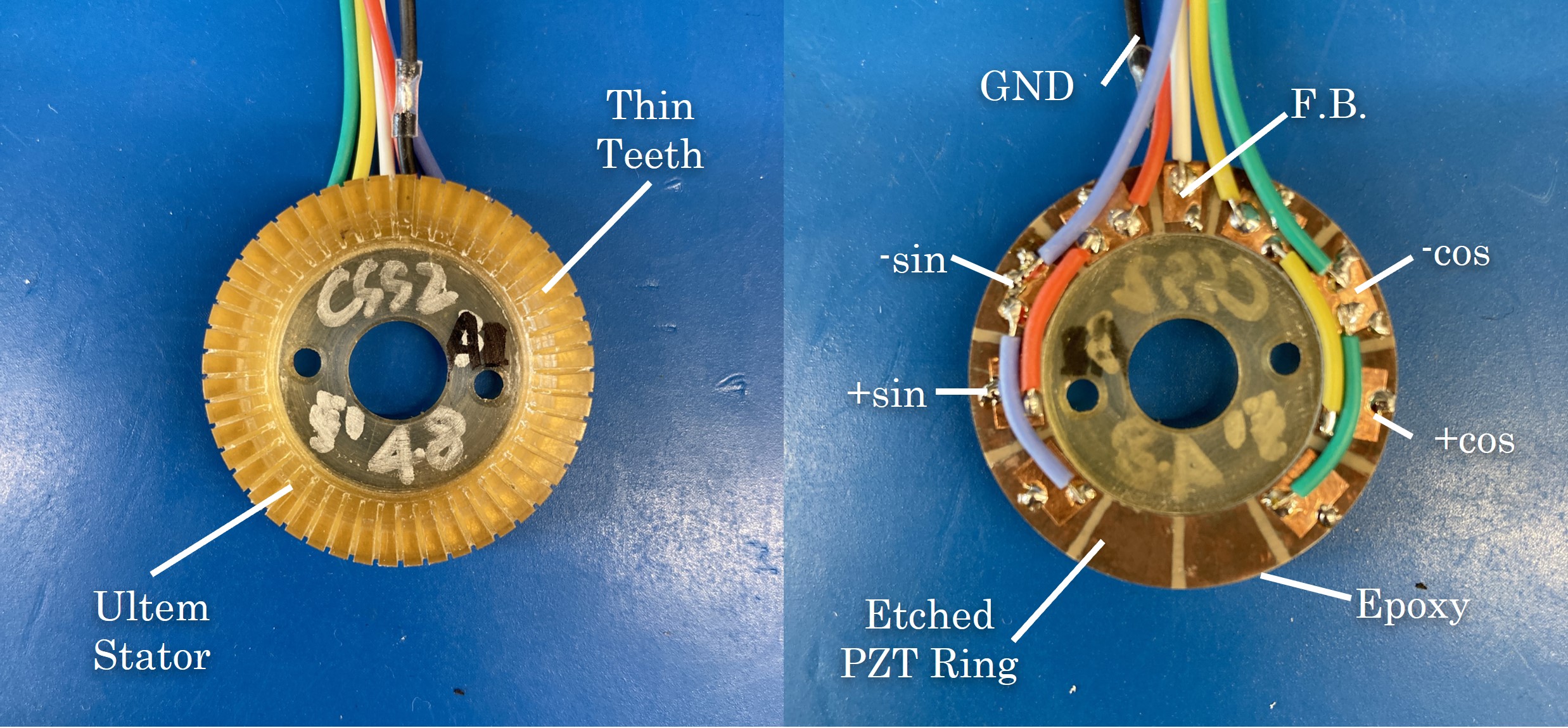}
    \caption{Detailed design of plastic stator assembly.}
    \label{fig:PMfinal}
\end{figure}

The CAD drawing of the MRI-compatible ultrasonic motor (USM) is depicted in Figure \ref{fig:PlasticDesign}, while the construction of the plastic USM is illustrated in Figure \ref{fig:PMpart} and \ref{fig:PMassem}, showcasing the individual components and the enclosed motor. The motor comprises a stator assembly, a rotor, and an output shaft, all encased within a base and cover. The stator assembly features an Ultem plastic stator integrated with a PZT-5H ceramic ring, secured at the bottom with conductive glue and fixed to the base center using a nylon nut. The finalized plastic stator is shown in Figure \ref{fig:PMfinal}. An etched PZT-5H ring was attached to the bottom of the optimized Ultem plastic stator using conductive glue, a detailed fabrication process can be found in \cite{zhao2021preliminary,dadkhah2016self,dadkhah2018increasing}. Several copper tape pieces were cut into specific-sized rectangles and applied to each electrode section, with AWG 22 wires soldered onto the tape to transmit 4-phase shifted signals, as soldering onto copper tape is much easier. Additional soldering was done between each tape and electrode to ensure better voltage signal propagation. The rotor makes frictional contact with the stator on its underside, while its topside connects to the output shaft via a rubber or foam ring, which provides pre-pressure and enhances frictional engagement with the rotor. A layer of Teflon, serving as a tribological material, is applied to the rotor side that directly contacts the stator. The shaft end is supported by a nylon-caged glass ball bearing, seated into the casing, with the output side of the shaft extending through the base's center hole. The base and cover enclose all components, secured with four nylon screws. Notably, the base, cover, rotor, and shaft were all 3D-printed using a Formlabs 2 (MA, USA) printer with acrylic-like resin. Thus, aside from the wires and electrode plating on the PZT crystal, all components are MRI-compatible.

\subsection{MRI-Compatible High Voltage USM Driver}

\begin{figure}
    \centering
    \includegraphics[width=0.9\linewidth]{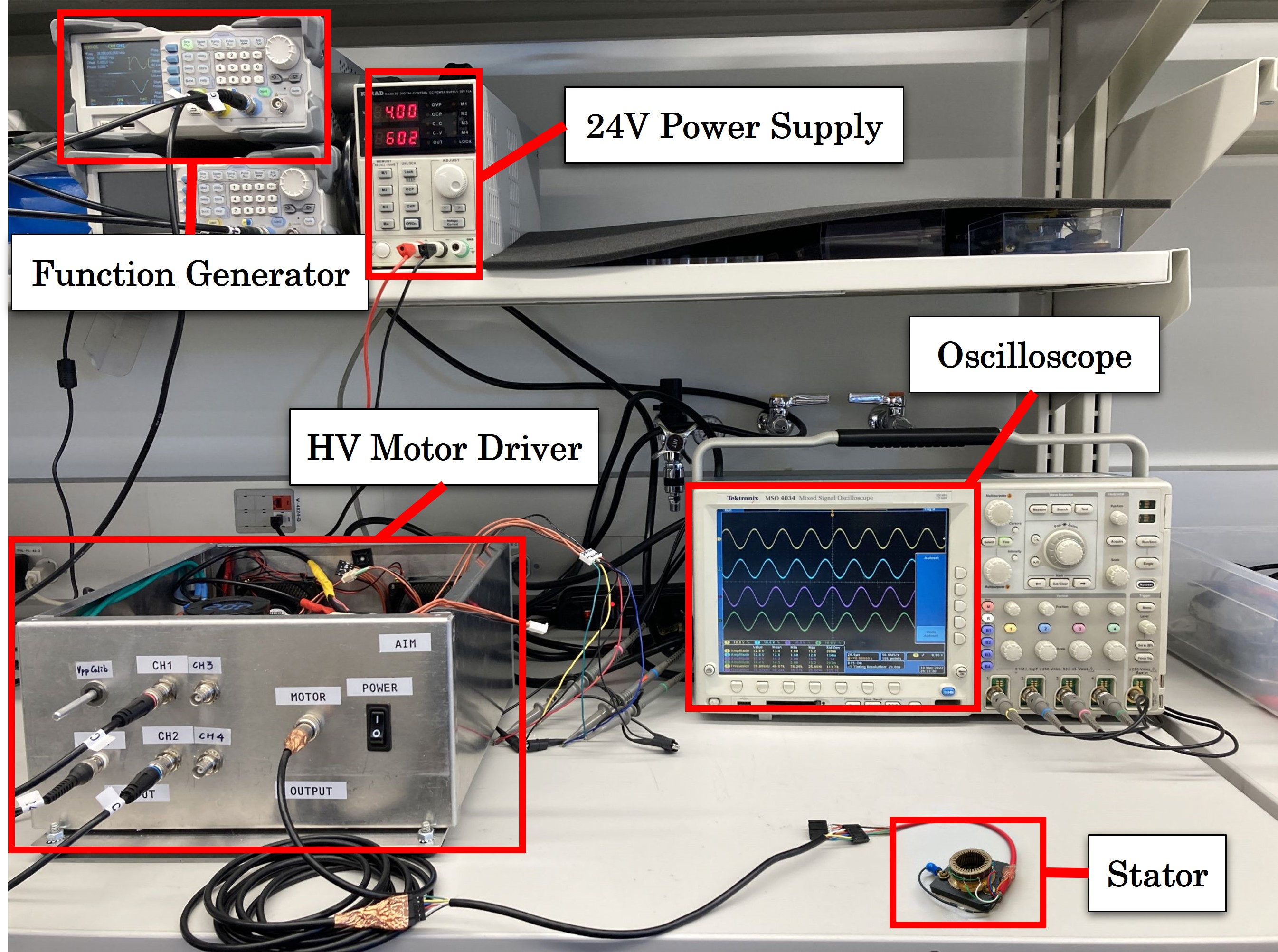}
    \caption{A remake high voltage USM driving system was configured. This system consists of an MRI-compatible HV motor driver, a 24V power supply, and a function generator. An oscilloscope is connected to the output testing pin for monitoring the signals.}
    \label{fig:HVdriverSetup}
\end{figure}

A high-voltage USM driving system was established for this study, comprising several key components: an HV motor driver, a 24V power supply, and a waveform generator. The 24V power supply provided power to the motor driver, while the waveform generator (Rigol\textregistered~DG1022Z, China) produced two sine signals, phase-shifted by 90$^\circ$, as the main inputs. Both the 24V power and input signals were connected to the BNC connectors on the front of the motor driver, with a 6-pin output cable used to connect to the motor. Multiple BNC cables were designed to transfer different types of signals, allowing for easy connection and power/signal transmission through the penetration panel on the wall between the MRI suite and the MRI console. An oscilloscope was connected to the internal output signal testing pins to monitor the signals applied to the stator. The driving system setup is shown in Figure \ref{fig:HVdriverSetup}.

\begin{figure}
    \centering
    \includegraphics[width=0.8\linewidth]{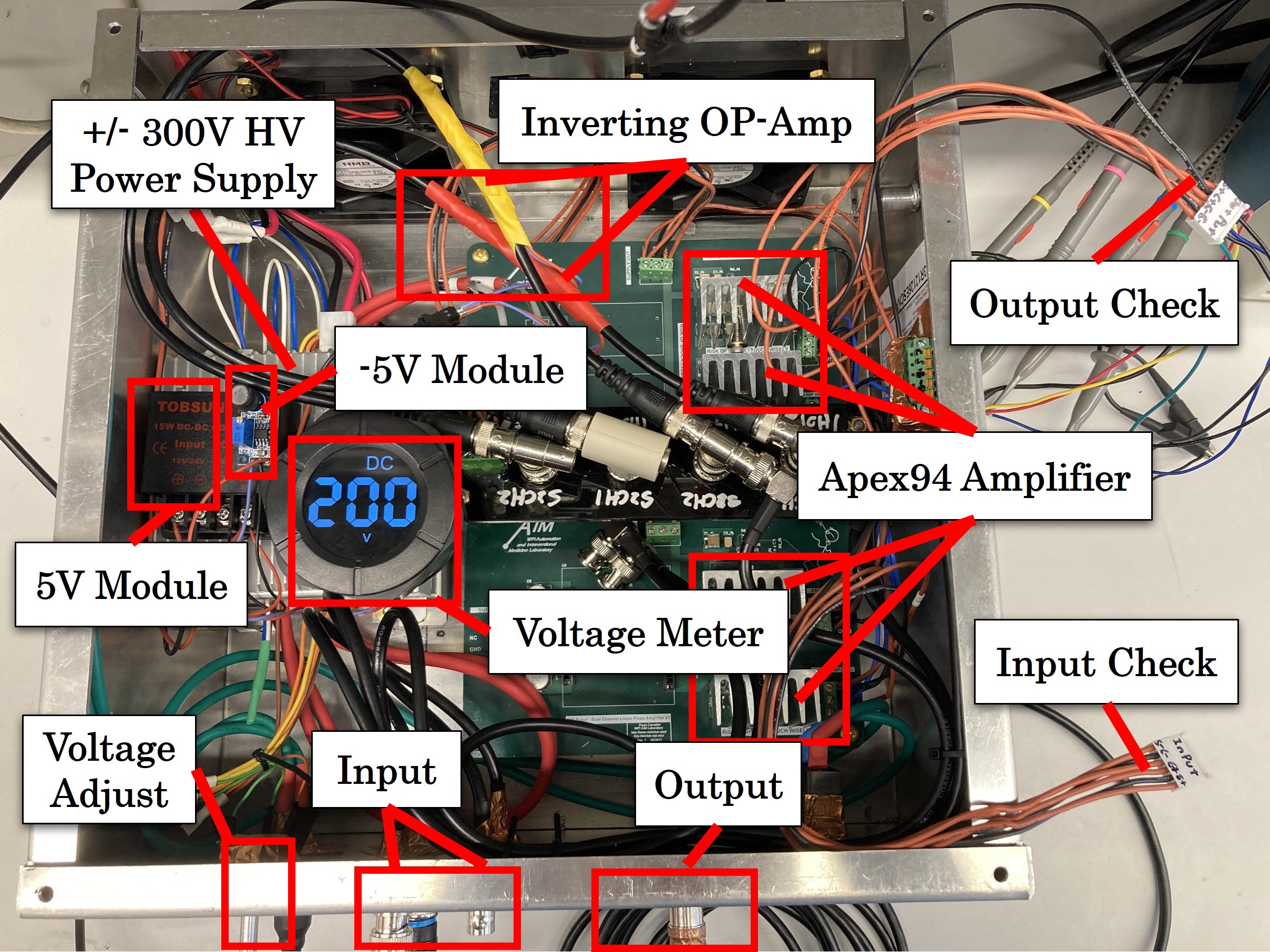}
    \caption{Construction of MRI-compatible HV motor driver box.}
    \label{fig:HVdriver}
\end{figure}

Figure \ref{fig:HVdriver} shows the HV motor driver, designed as a universal driving and testing system compatible with MRI environments. All components are enclosed within an aluminum box to contain signals and minimize electromagnetic interference (EMI) with the MRI scanner. The 24V power supply energizes a pair of RC125-0.3P and RC125-0.3N (Matsusada\textregistered, Japan) DC-DC 300V HV power supplies, as well as a 5V power module. The 5V module connects to a voltage inverter to generate -5V, effectively creating a -5V power module. A potentiometer controls the HV DC-DC power supply to adjust the paired maximum voltage outputs, while a voltage meter in the center displays the voltage output—note that only the +300V power supply voltage is shown.

\begin{figure}
    \centering
    \includegraphics[width=0.9\linewidth]{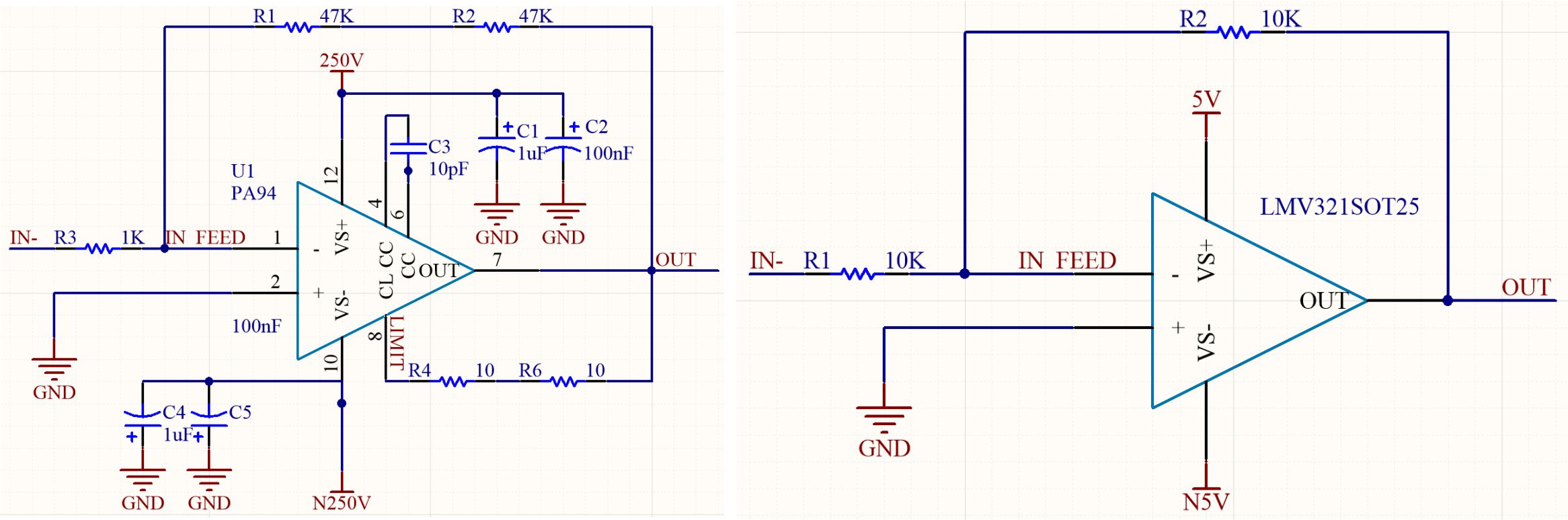}
    \caption{Amplifier schematic. (Left) PA94-based 100$\times$ (specifically based on the resistor ratio calculation it is 94$\times$ amplifying) inverting amplifier. (Right) LMV321SOT25-based frequency synchronizing reversing inverting amplifier.}
    \label{fig:amplifier}
\end{figure}

The driver uses two dual-channel amplifier PCBs. The 90° phase-shifted signals generated by the function generator, labeled as \emph{sin} and \emph{cos}, are directly connected to the first dual-channel amplifier PCB as the initial positive input signals. Two custom-made signal-inverting cables, employing the LMV321SOT25 (Diodes Incorporated, USA) and following the schematic shown in the right figure of Figure \ref{fig:amplifier}, invert the \emph{sin} and \emph{cos} signals into \emph{-sin} and \emph{-cos}. These inverted signals are then fed into the second dual-channel amplifier PCB, generating a 4-phase input power. The resulting four signals are amplified by PA94 (Apex Microtechnology, USA) operational amplifiers, configured in an inverting setup and powered by the HV DC-DC power supplies.

\begin{figure}
    \centering
    \includegraphics[width=1\linewidth]{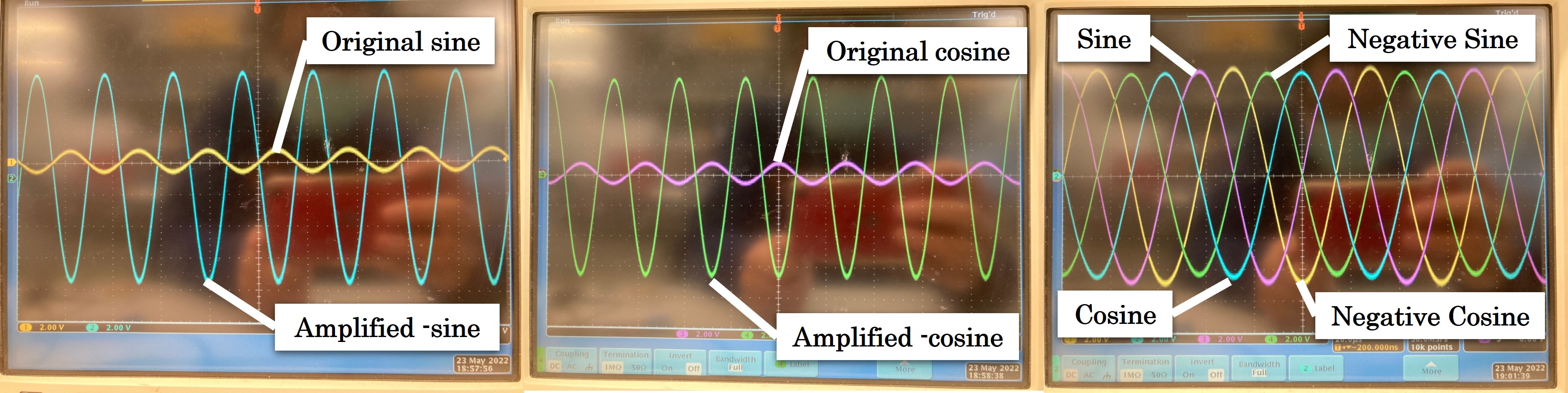}
    \caption{Signals amplify process and compare. (Left) \emph{sin} signal inverted and amplified to 100$\times$ \emph{-sin}. (Middle) \emph{cos} signal inverted and amplified to 100$\times$ \emph{-cos}. (Right) Final 4-phase amplified signals output without being connected to a motor.}
    \label{fig:signal}
\end{figure}

The output signals without connecting to a motor are shown in Figure \ref{fig:signal}. PA94-based inverting amplifiers and the paired HV power supplies were used to generate inverted amplified power, namely from \emph{sin} to 100x amplified \emph{-sin}, and another three signals can be generated accordingly.  Note that the amplitude used on the waveform generator can not exceed the voltage limitation from paired HV DC-DC power supplies after amplified, otherwise, it will not work functionally, and the signal wave will have an abnormal shape, and finally will cause the motor to nonfunctional and even brake by the abnormal input signals.

This driving system enables us to operate our custom-made USMs for testing, validation, and parameter measurement. Additionally, the box can serve as an auxiliary device for the development of new auto-tuning USM driver PCB boards.

\subsection{Motor Output Parameters Testing Platform}

\begin{figure}
    \centering
    \includegraphics[width=1\linewidth]{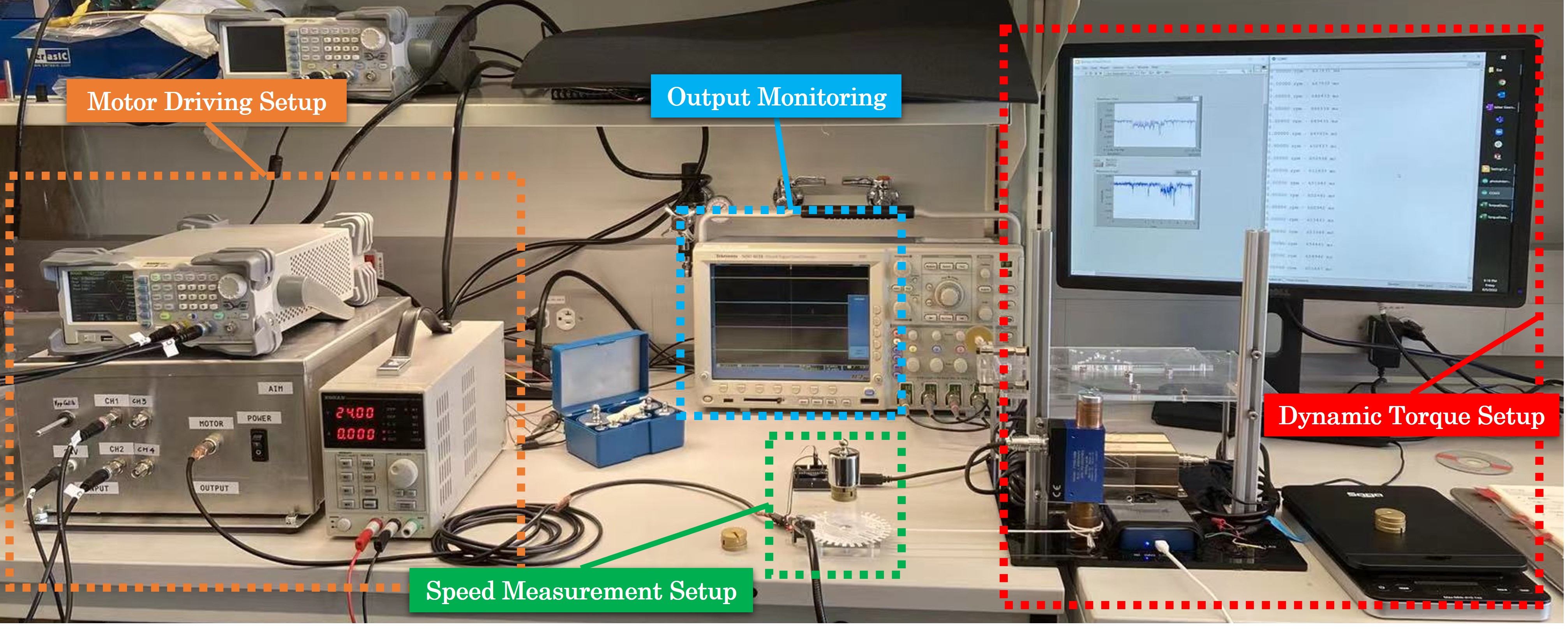}
    \caption{Testing platform setup, including HV motor driver, speed measurement system, dynamic torque measurement system, and a signal output monitor.}
    \label{fig:testplatformall}
\end{figure}

A testing platform including rotary speed and dynamic torque measurement functionality was built and used for measuring the plastic motor output performance. As shown in Figure \ref{fig:testplatformall}, this system consisted of multiple modules, namely the motor driving setup, speed measurement setup, dynamic torque setup, and output monitoring. The oscilloscope was used as a motor output signals monitoring device for confirming the signal output functionality.

\subsubsection{Dynamic Torque Output Measurement}

The dynamic torque measurement system is depicted on the left side of Figure \ref{fig:measurement}. It consisted of a torque meter system, a pulling string connecting the torque sensor to the rotor, and multiple weights applied to the motor assembly with different testing parameters. The weights were measured using scales before testing. The torque sensor (FY02-1NM, Forsentek\textregistered, China) was mounted vertically, with no load applied to the output shaft side. Torque data was collected by the torque meter system and later post-processed.

\begin{figure}
    \centering
    \includegraphics[width=0.9\linewidth]{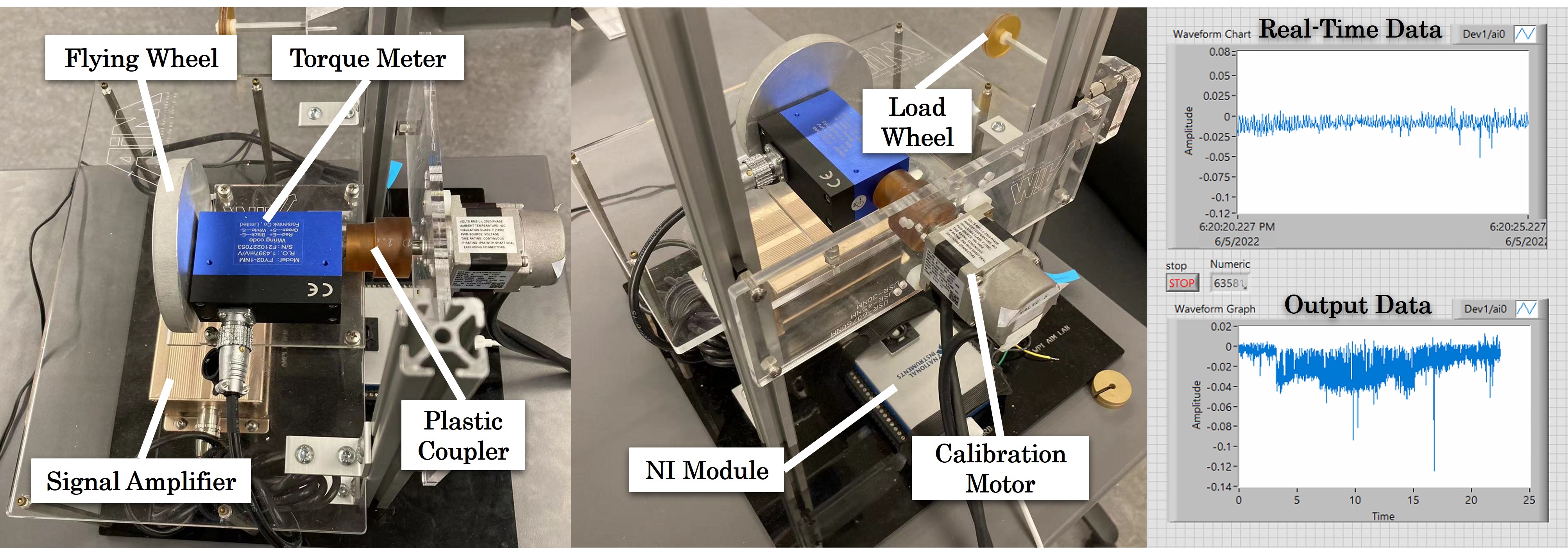}
    \caption{Testing setup of torque meter.}
    \label{fig:A_TorqueMeter}
\end{figure}

\begin{figure}
    \centering
    \includegraphics[width=1\linewidth]{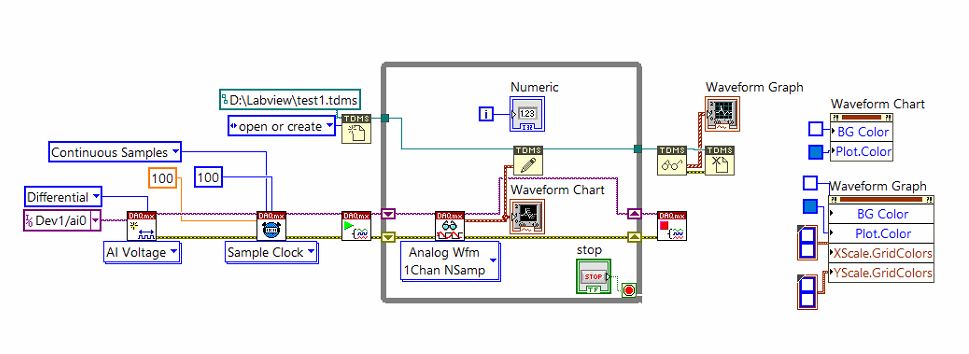}
    \caption{LabView code for USB-6001 Daq card.}
    \label{fig:A_labview}
\end{figure}

The testing setup is shown in Figure \ref{fig:A_TorqueMeter}. The torque sensor was fixed to the frame, with its input connected to a testing calibration motor via a plastic coupler to minimize signal interference. The output side was connected to an aluminum flywheel. The sensor signal cable was linked to a signal amplifier, which was then connected to an NI USB-6001 DAQ card for data acquisition. A LabView program, shown in Figure \ref{fig:A_labview}, was developed to control the NI module through a user interface (UI), which featured a real-time data window and an output data window. A load wheel was used to measure the actual motor torque under load.

\subsubsection{Rotary Speed Measurement}

\begin{figure}
    \centering
    \includegraphics[width=0.9\linewidth]{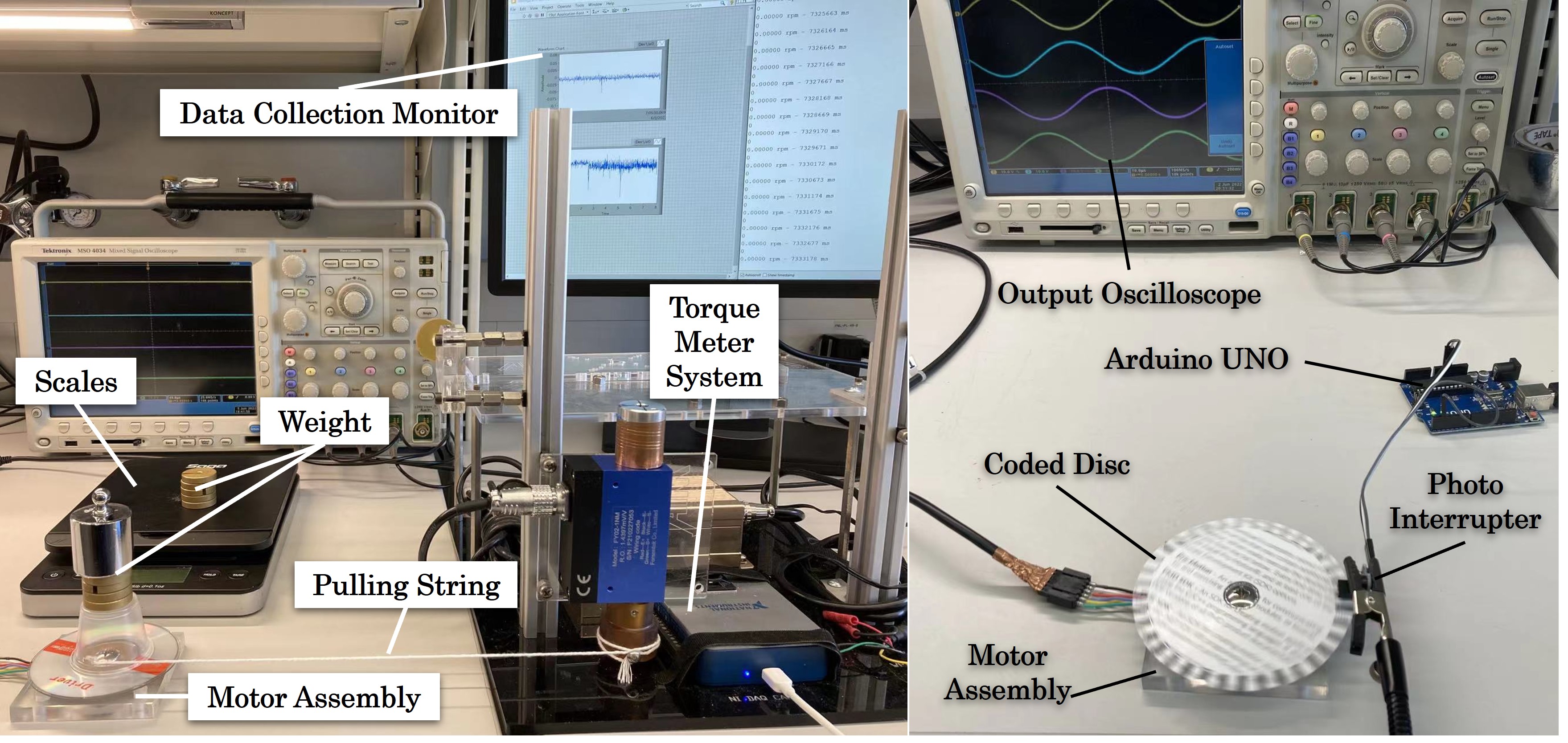}
    \caption{(Left) Torque measurement system. (Right) Rotary speed measurement system.}
    \label{fig:measurement}
\end{figure}

The rotary speed measurement system was built up and can be found in Figure \ref{fig:measurement} right figure. This system consisted of a photo-interrupter, an Arduino UNO for collecting the interrupter data, a hard paper-made coded disc with 36 teeth equally along the circumference fixed onto the CD disc acting as rotor, and the output oscilloscope was a monitor for motor output signals to monitor. When the coded disc teeth were in the middle of the gantry of the photo-interrupter, the infrared shooting to the transistor drained voltage drops, then Arduino UNO collected the data and connected it to the computer to display and save the rotary speed data from the serial monitor showing on the monitor. 

\section{Method}

\subsection{Signal-to-Noise Ratio (SNR) Measurement}

\begin{figure}
    \centering
    \includegraphics[width=1\linewidth]{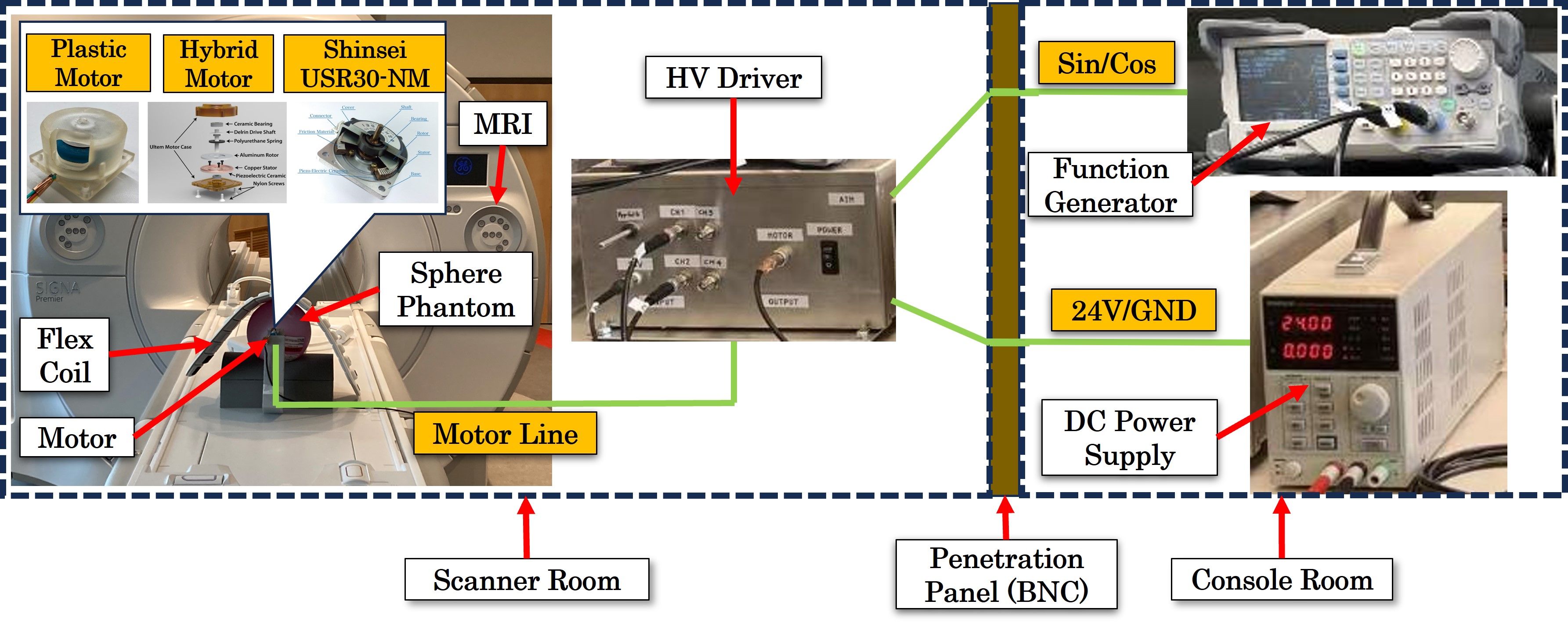}
    \caption{Noise listening setup. The function generator and DC power supply were placed in the MRI console room, both devices used BNC cables to input signal and 24V power into the motor driver through a penetration panel. The driver was deployed at the corner of the MRI room to reduce the interference. An SNR sphere phantom was located on the MRI bed, and 3 types of motors were placed at the center side surface of the phantom. A Flex coil was covered onto the phantom and motor setup.}
    \label{fig:SNRsetup}
\end{figure}

The complete plastic motors were analyzed inside a GE Signa 3T scanner. An MRI SNR phantom was used as the object being imaged on the MRI table, testing motors were placed on the middle height and center of the sphere phantom surface, then covered with GEM FLEX COIL (GE, USA) MRI Coil, a noise listening setup can be found in Figure \ref{fig:SNRsetup}. Motors were driven by the HV USM driver system placed at the corner of the MRI room. Multiple conditions were scanned by MRI, which is shown in Table \ref{tbl:condition}. The Hybrid Motor indicates the plastic-enclosed USR60-based plastic motor developed by Carvalho \textit{et al.} in \cite{carvalho2020demonstration}, which was also compared in this study. For each of these conditions, image quality was assessed using two imaging sequences, namely T1 and T2-weighted sequences Fast Spin Echo (FSE), and the parameters used for these scans can be found in Table \ref{tbl:sequence}. 

\begin{table}
    \caption{SNR scan conditions description and condition name.}
    \vspace{1mm}
    \centering
    \begin{tabular}{ll}
    \hline
    Condition Description     & Condition Name   \\ \hline\hline
    Phantom Only              & Baseline         \\ \hline
    HV Driver Box On          & Driver ON         \\ \hline
    Plastic Motor 1 Stop      & PM1NotMoving     \\ \hline
    Plastic Motor 1 Spinning  & PM1Moving        \\ \hline
    Plastic Motor 2 Stop      & PM2NotMoving     \\ \hline
    Plastic Motor 2 Spinning  & PM2Moving        \\ \hline
    Hybrid Motor Stop         & HybridNotMoving  \\ \hline
    Hybrid Motor Spinning     & HybridMoving     \\ \hline
    Shinsei USR30-NM Stop     & ShinseiNotMoving \\ \hline
    Shinsei USR30-NM Spinning & ShinseiMoving    \\ \hline
    \end{tabular}
    \label{tbl:condition}
\end{table}

\begin{table}[ht!]
    \caption{Image quality tests: Scan sequence parameters.}
    \vspace{1mm}
    \centering
   \resizebox{\linewidth}{!}{
   \begin{tabular}{|c|c|c|c|c|c|c|c|}
    \hline
    Sequance & \begin{tabular}[c]{@{}c@{}}TE\\ (ms)\end{tabular} & \begin{tabular}[c]{@{}c@{}}TR\\ (ms)\end{tabular} & \begin{tabular}[c]{@{}c@{}}Slice \\ Thickness \\ (mm)\end{tabular} & \begin{tabular}[c]{@{}c@{}}Bandwidth\\ (hz/pixel)\end{tabular} & \begin{tabular}[c]{@{}c@{}}Image Size\\ (pixels)\end{tabular} & \begin{tabular}[c]{@{}c@{}}FOV\\ (mm)\end{tabular}  & \begin{tabular}[c]{@{}c@{}}Slices\\ Quantity\end{tabular}  \\ \hline
    T1 FSE   & 20  & 500  & 3.0 & 278 & 256 & 200 & 8\\ \hline
    T2 FSE   & 55  & 2000   & 3.0   & 278  & 256   & 200  & 19\\ \hline
    \end{tabular}
    }
    \label{tbl:sequence}
\end{table}

\begin{figure}
    \centering
    \includegraphics[width=0.8\linewidth]{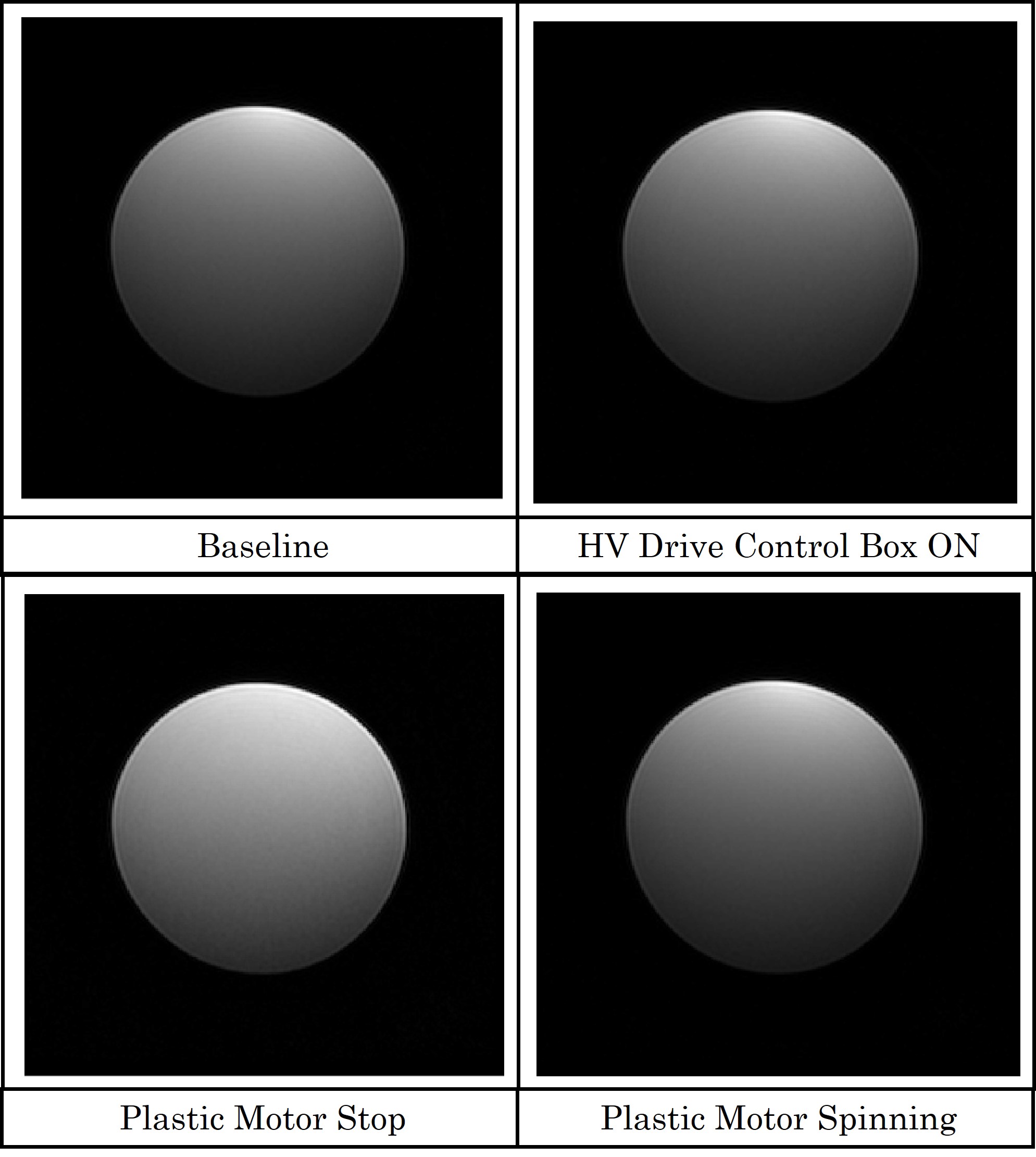}
    \caption{Selected T2 scanning images on 9$^{th}$ slice with baseline, HV driver on, plastic motor stop, and plastic motor spinning conditions.}
    \label{fig:SNRimage}
\end{figure}

Figure \ref{fig:SNRimage} shows the selected T2 scanning images with the same slice (slice \#9) under several conditions, namely Baseline, HV Driver Control Box On, Plastic Motor Stop, and Plastic Motor Spinning. The SNR drop was not significant enough to be apparent upon visual inspection of the images under different conditions as can be observed. 

\begin{figure}
    \centering
    \includegraphics[width=0.5\linewidth]{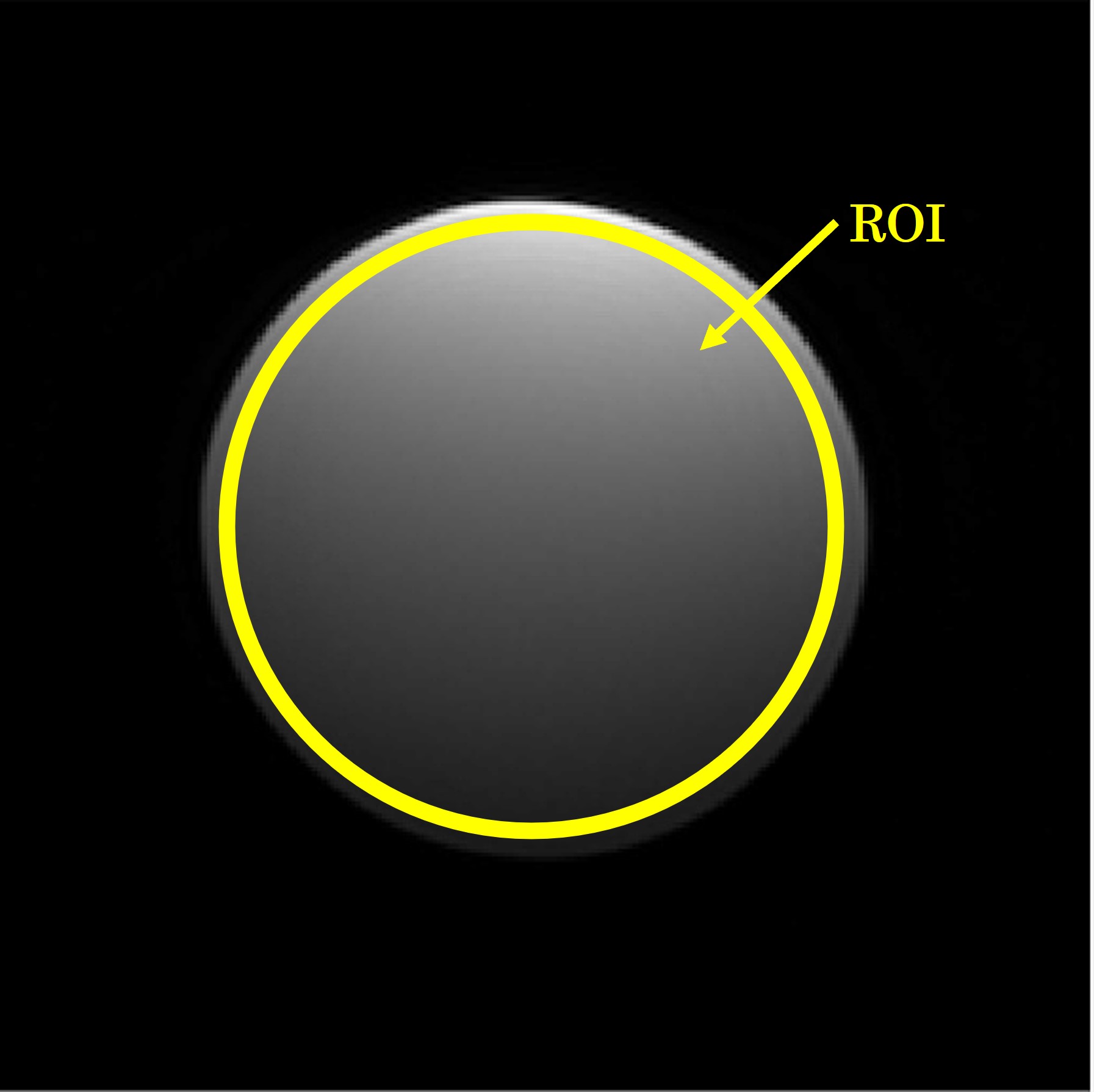}
    \caption{Selected T2 scanning images on 9$^{th}$ slice with baseline, HV driver on, plastic motor stop, and plastic motor spinning conditions.}
    \label{fig:ROI}
\end{figure}

SNR was calculated using the image difference method described in \cite{NEMA2008_SNR}, referred to as ``method 1". A circular region of interest (ROI), shown in Figure \ref{fig:ROI}, was selected to measure the mean signal $\mu$. This ROI, which encompassed most of the phantom, was kept constant across all image sequences. Initially, a scan was performed with only the phantom inside the MRI room, without the HV driver present; this was designated as "baseline 0". Following the conditions outlined in Table \ref{tbl:condition}, scans were conducted for all 10 conditions. Each of these 10 images was subtracted from the initial "baseline 0" image for each slice pair to produce new images. The standard deviation of pixel values within the ROI of each subtracted image provided an estimate of the random noise, which was assumed to be Gaussian. Since this calculation involves a difference operation, the noise $\sigma$ was corrected by a factor of $1/\sqrt{2}$. The SNR was then defined as shown in Equation \ref{equ:SNR}. SNR values across all slices of the same region were compared to the baseline condition using a paired t-test in R\textregistered~software to determine if there was a statistically significant change in the mean.

\begin{equation}
    SNR = \frac{\mu}{\sigma}
    \label{equ:SNR}
\end{equation}

\subsection{Stator Parameter Influence on Motor Speed and Torque Performance}

\begin{figure}
    \centering
    \includegraphics[width=0.6\linewidth]{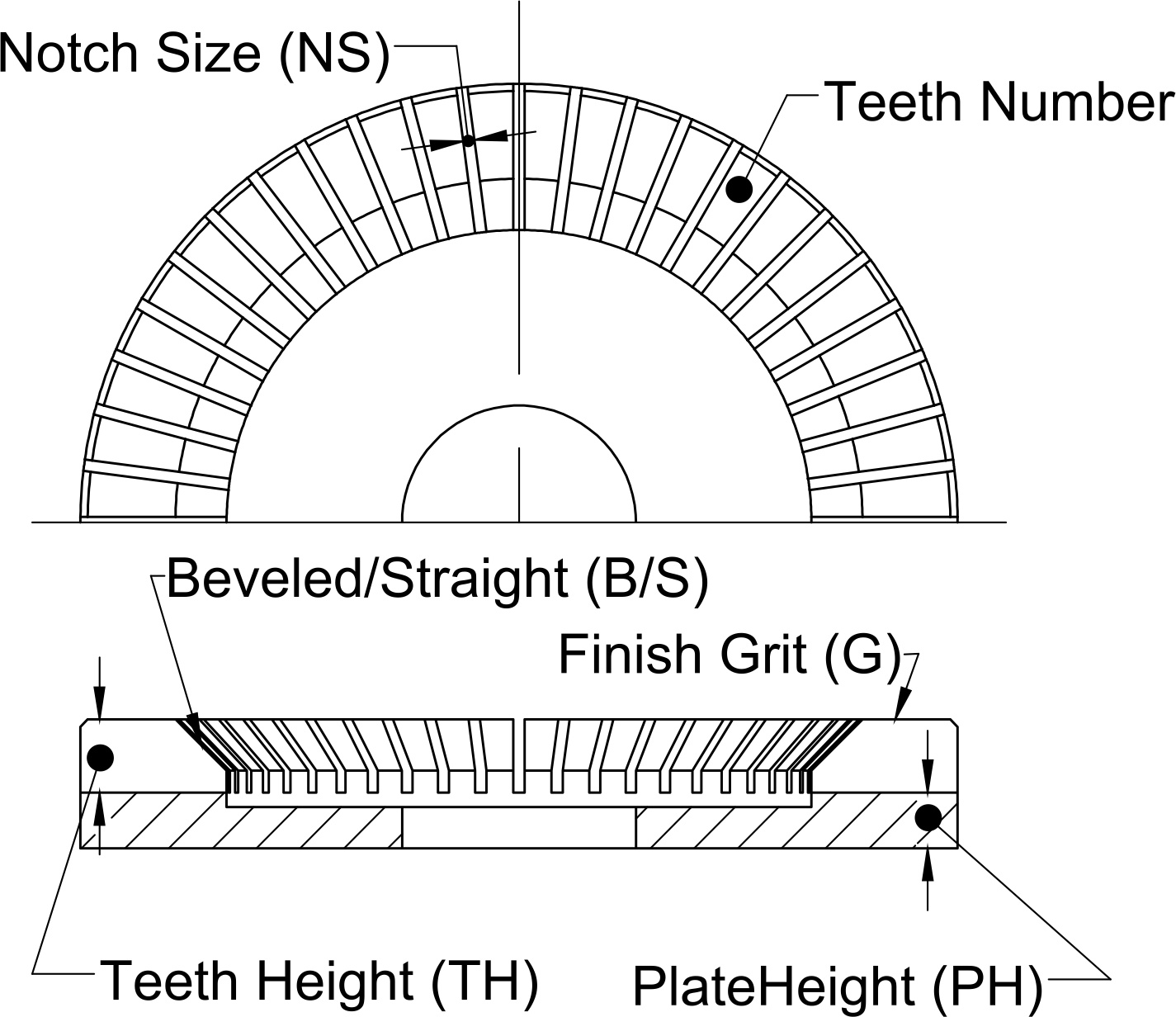}
    \caption{Drawing of stator parameters discussed in this section. BL - baseline, SE - straight edge, WN - wide notch, G - sandpaper grit number.}
    \label{fig:parameter}
\end{figure}

Multiple plastic stators with different parameters were fabricated, and all the parameter variations are listed in Table \ref{tbl:motorparameters}, namely, teeth number (T\#), teeth height (TH), plate height (PH), notch size (NS), finish grit (G), and edge bevel/straight (B/S), all these parameters discussed in this section can be found in Figure \ref{fig:parameter}. In this study, five groups of plastic stators were fabricated, namely, groups A through E. Group A was designed as an optimized stator baseline, with 48 teeth and 0.39mm notch size, used grit\#1000 sandpaper for surface finishing, and beveled edge. Group B and C were compared with A, while Group D was compared with B, and Group E was compared within the group. Group B was made with 40 teeth, group C was made with straight, or non-beveled edges, group D was made with a wider notch of 0.75mm, and Group E used different grit number of sandpaper to achieve different levels of surface roughness, namely grit of 100, 1000, 5000, and 10000. Note that the E1 stator was another equivalent stator of group A baseline plastic stator. Based on these groups of the stator, multiple rotary speed and output torque versus the prepressure applied on the motor were studied.

\begin{table}
    \caption{All the plastic stators parameters description. BL - baseline, SE - straight edge, WN - wide notch, G - sandpaper grit number.}
    \vspace{1mm}
    \centering
    \resizebox{\linewidth}{!}{
   \begin{tabular}{|c|c|c|c|c|c|c|}
    \hline
    Stator\# & \begin{tabular}[c]{@{}c@{}}Teeth\#\\ (T\#)\end{tabular} & \begin{tabular}[c]{@{}c@{}}Teeth Height\\ (TH)(mm)\end{tabular} & \begin{tabular}[c]{@{}c@{}}Plate Height \\ (PH)(mm)\end{tabular} & \begin{tabular}[c]{@{}c@{}}Notch Size\\ (NS)(mm)\end{tabular} & \begin{tabular}[c]{@{}c@{}}Finish Grit\\ (G)\end{tabular} & \begin{tabular}[c]{@{}c@{}}Edge Bevel\\ (B/S)\end{tabular} \\ \hline
    A1, BL   & 48  & 2.50  & 1.74 & 0.39 & 1000 & Beveled \\ \hline
    A2, BL   & 48  & 2.73  & 1.75 & 0.39 & 1000 & Beveled \\ \hline
    B1, 40T   & 40  & 2.52  & 1.74 & 0.39 & 1000 & Beveled \\ \hline
    B2, 40T   & 40  & 2.79  & 1.70 & 0.39 & 1000 & Beveled \\ \hline
    C1, SE   & 48  & 2.48  & 1.75 & 0.39 & 1000 & Straight \\ \hline
    C2, SE   & 48  & 2.72  & 1.70 & 0.39 & 1000 & Straight \\ \hline
    D1, WN   & 40  & 2.48  & 1.81 & 0.75 & 1000 & Beveled \\ \hline
    D2, WN   & 40  & 2.79  & 1.69 & 0.75 & 1000 & Beveled \\ \hline
    E0, G100   & 48  & 2.50  & 1.89 & 0.39 & 100 & Beveled \\ \hline
    E1, G1000   & 48  & 2.56  & 1.91 & 0.39 & 1000 & Beveled \\ \hline
    E2, G5000   & 48  & 2.54  & 1.87 & 0.39 & 5000 & Beveled \\ \hline
    E3, G10000   & 48  & 2.62  & 1.83 & 0.39 & 10000 & Beveled \\ \hline
    \end{tabular}
 }
    \label{tbl:motorparameters}
\end{table}

\section{Results}

\subsection{SNR Drop Result}
\begin{figure}
    \centering
    \includegraphics[width=0.9\linewidth]{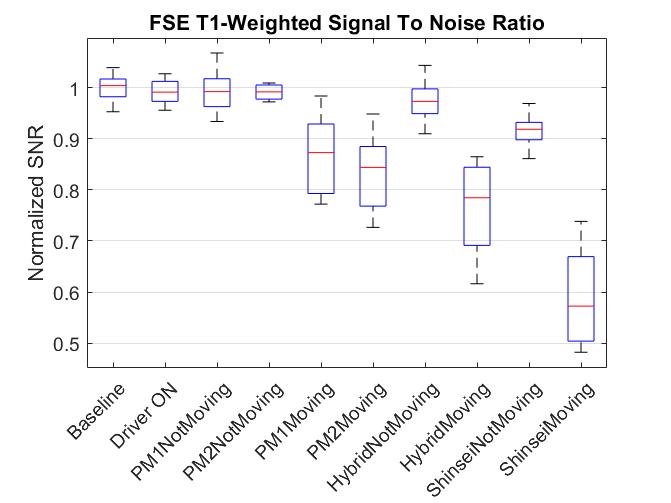}
    \caption{T1 weighted SNR results of the nine different testing conditions. For each condition, SNR was calculated on 8 slices through the homogeneous region of the phantom. Results are normalized with respect to the baseline scan.}
    \label{fig:T1result}
\end{figure}

\begin{figure}
    \centering
    \includegraphics[width=0.9\linewidth]{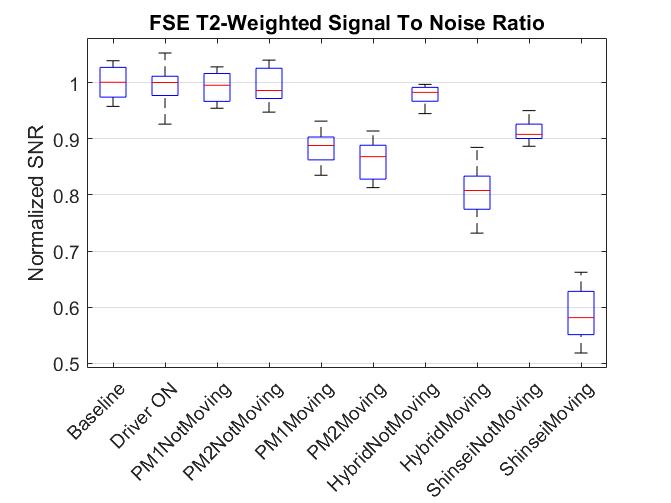}
    \caption{T2 weighted SNR results of the nine different testing conditions. For each condition, SNR was calculated on 19 slices through the homogeneous region of the phantom. Results are normalized with respect to the baseline scan.}
    \label{fig:T2result}
\end{figure}

Figure \ref{fig:T1result} and Figure \ref{fig:T2result} display the normalized SNR results relative to the baseline scan. The results indicate the following: (1) There is no statistically significant difference between the baseline and the driver ON conditions in both T1 and T2-weighted scans (p=0.29 and p=0.65, respectively), with the reduction in mean SNR being less than 1\%. (2) No statistically significant differences were found between the baseline and the non-moving conditions of both plastic motors (PM1 and PM2) in T1 and T2-weighted scans (p=0.7739, p=0.3491, p=0.5661, and p=0.5184 for T1 PM1, T2 PM1, T1 PM2, and T2 PM2 configurations, respectively), with SNR drops of less than 2\%. (3) There were low mean SNR drops for the hybrid motor and Shinsei USR30 in the non-moving condition (within 3\% and 9\%, respectively). The larger SNR drop for the USR30 motor in the non-moving condition may be due to distortion in the magnetic field near the imaging area, leading to a signal drop. (4) A significant statistical difference was observed for both plastic motors under spinning conditions (p<0.00 for all conditions and scans), with reductions in mean SNR reaching up to 13\% and 16\% for PM1 and PM2, respectively. However, the SNR drop for both plastic motors was smaller compared to the hybrid motor and USR30 motor, which experienced drops of up to 22\% and 43\%, respectively. This led to the development of a new plastic motor with greater MRI compatibility compared to the plastic-enclosed motor developed by Carvalho \textit{et al.} \cite{carvalho2020demonstration}.

\subsection{Motor Speed and Torque Performance}

\subsubsection{Group A: Baseline}

\begin{figure}
    \centering
    \includegraphics[width=0.8\linewidth]{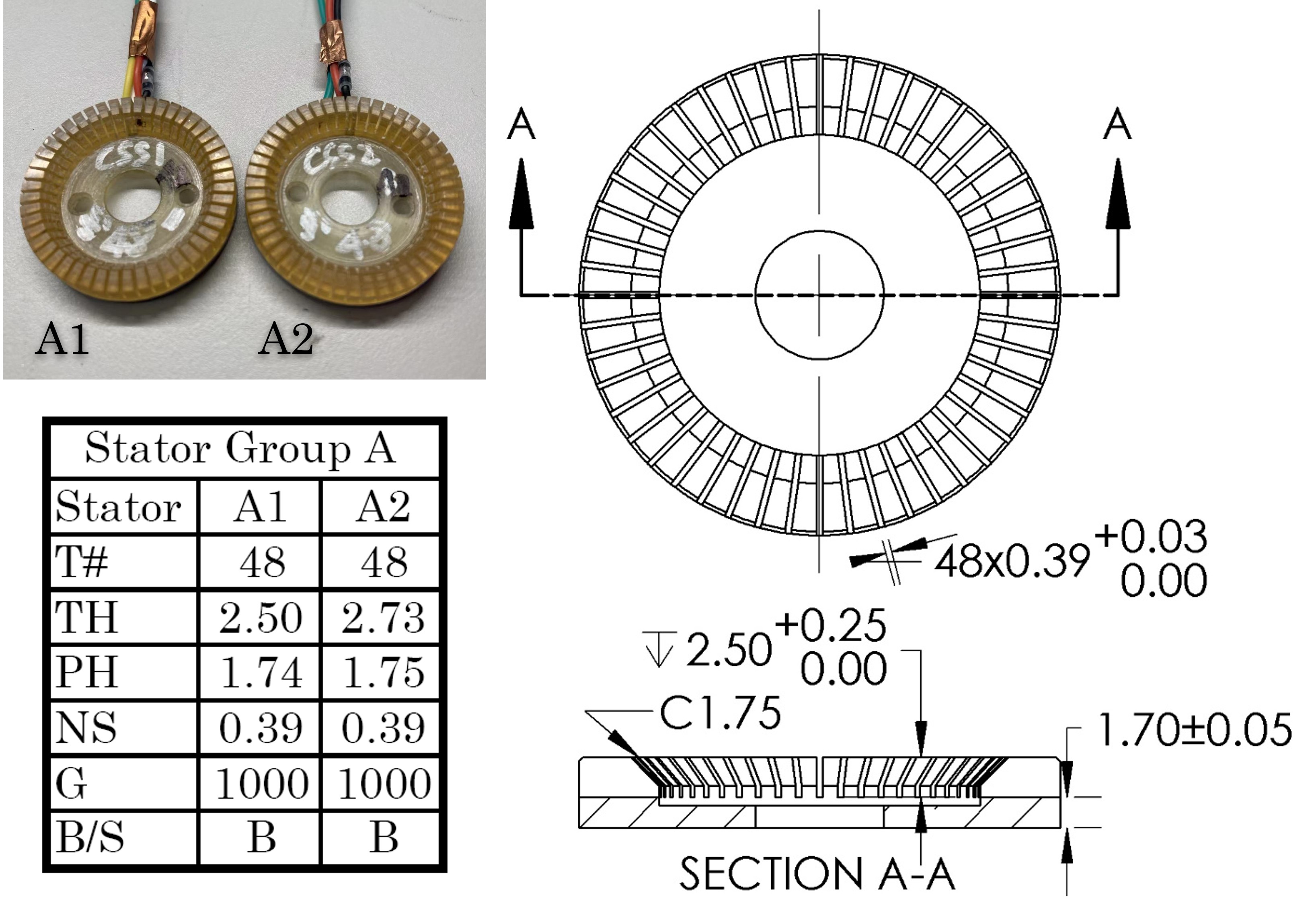}
    \caption{Group A plastic stators condition, including image, dimensional drawing, and parameters.}
    \label{fig:PMA}
\end{figure}

\begin{figure}
    \centering
    \includegraphics[width=0.8\linewidth]{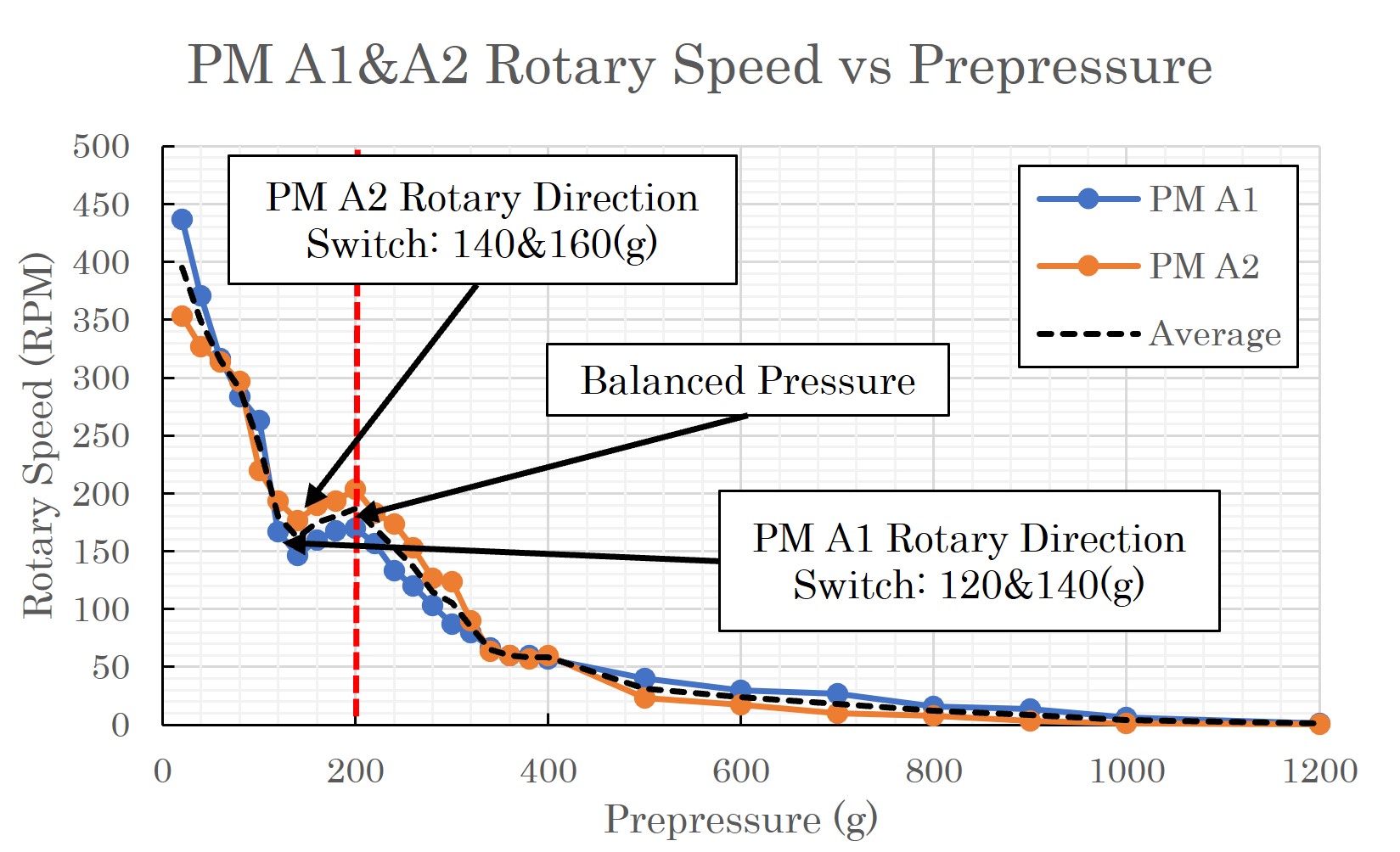}
    \caption{Group A motors rotary speed results versus prepressure.}
    \label{fig:PMAspeed}
\end{figure}

The condition of baseline group A can be found in Figure \ref{fig:PMA}, with image, dimensional drawing, and parameters. Figure \ref{fig:PMAspeed} shows the two plastic stators A1 and A2 rotary speed results versus prepressure, with 20g incremental from 0 to 400g and 100g incremental from 400 to 1200g, and collected the stable spinning data over 10s. Results show that the highest speed performed at the initial prepressure was 20g, where A1 reached 436.6665rpm and A2 reached 353.3333rpm stable rotary speed. When the prepressure reached 120g and 140g with A1 and A2 respectively, the speed reached a low point at 146.6667rpm and 176.6667rpm, next prepressure level both motors changed the rotation direction and the speed increased a little bit to 156.6667rpm and 183.3333rpm respectively, both motors reached the second high-speed point at approximately 200g, which was considered as a balance points over 200g the motors started to slow down because of the weight was increased and prevented the motor from spinning at high speed. Prepressure increased up to 1200g, at this point it was hard to measure the specific speed since the motor spun slowly and discontinuously. 

\begin{figure}
    \centering
    \includegraphics[width=0.8\linewidth]{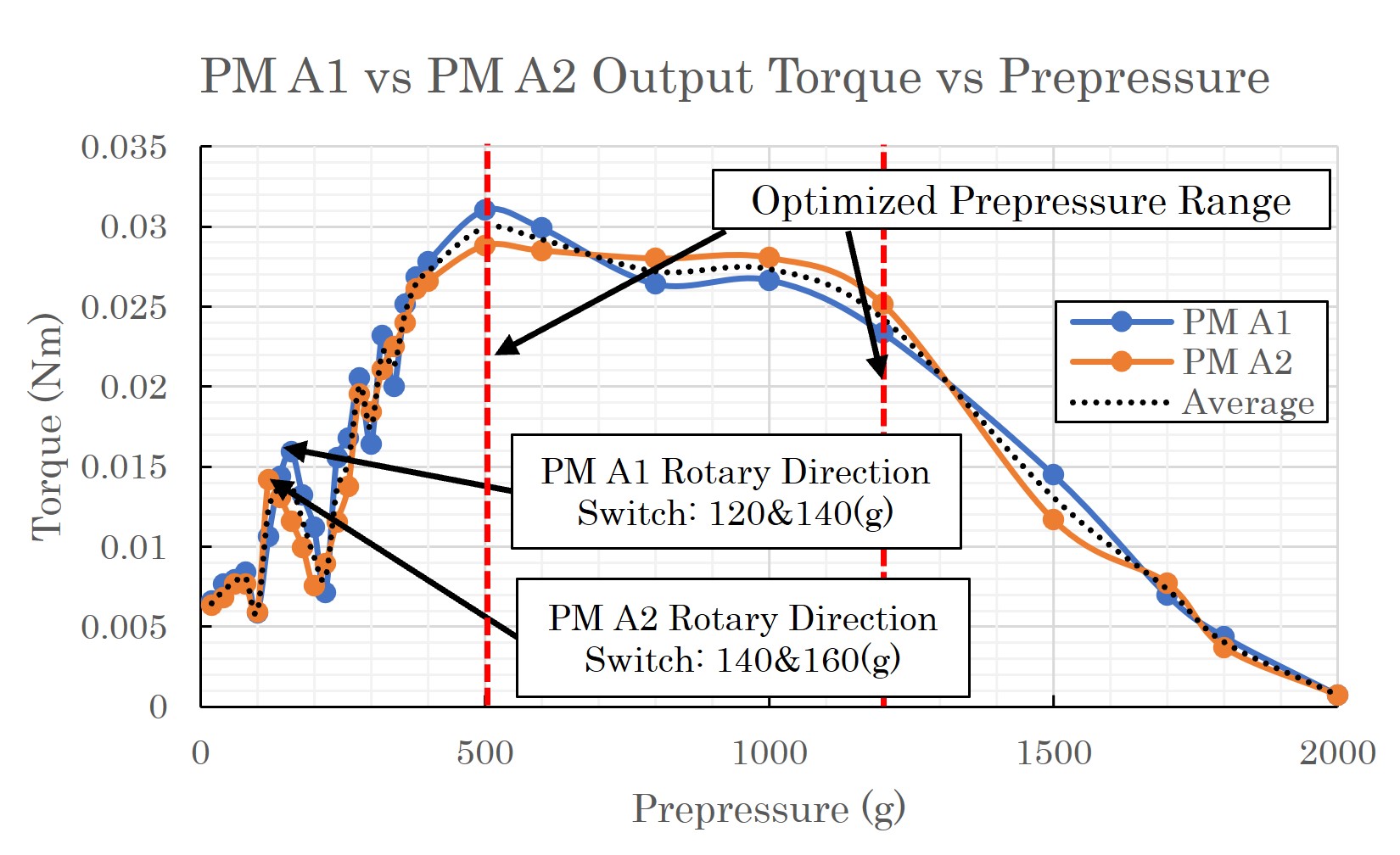}
    \caption{Group A motors output torque results versus prepressure.}
    \label{fig:PMAtorque}
\end{figure}

 Figure \ref{fig:PMAtorque} shows the average output torque result under each prepressure value. Similar to speed measurement, the prepressure increased with 20g incremental from 0 to 400g and 100g incremental from 400 to 1200g, five trials were measured and tested. The results show a very close average output torque performance between A1 and A2, and the largest torque output performed at approximately 500g prepressure, where A1 provided 0.0348Nm and A2 provided 0.0304Nm. Considering the speed measurement data, the torque value also performed a high point and then a drop at 120g and 140g with A1 and A2 respectively. When the prepressure reached approximately 200g, the torque value reached a low point and the next pressure increased. Because the speed and torque of the motor are inversely proportional, so torque data matched with speed value change. Although the highest torque was performed when prepressure reached 500g, from 500g to 1200g the torque remained high output and changed very small, however when the prepressure was above 1200g, torque output decreased again. Until 2000g both motors still had output, with a very low-value approximately 0.0005Nm and 0.0007Nm respectively.

\subsubsection{Group B: Teeth Number}

\begin{figure}
    \centering
    \includegraphics[width=0.8\linewidth]{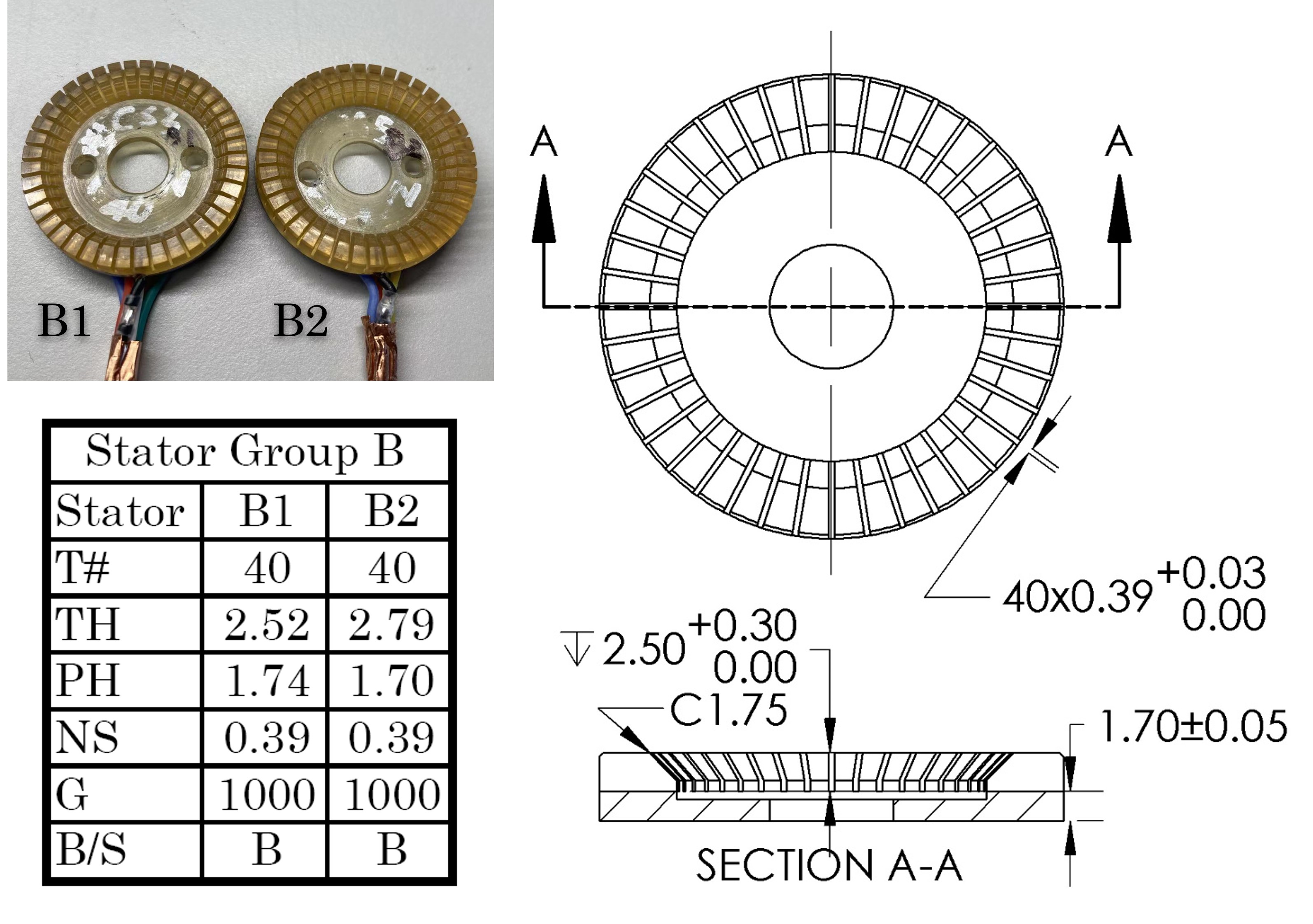}
    \caption{Group B plastic stators condition, including image, dimensional drawing, and parameters.}
    \label{fig:PMB}
\end{figure}

\begin{figure}
    \centering
    \includegraphics[width=0.8\linewidth]{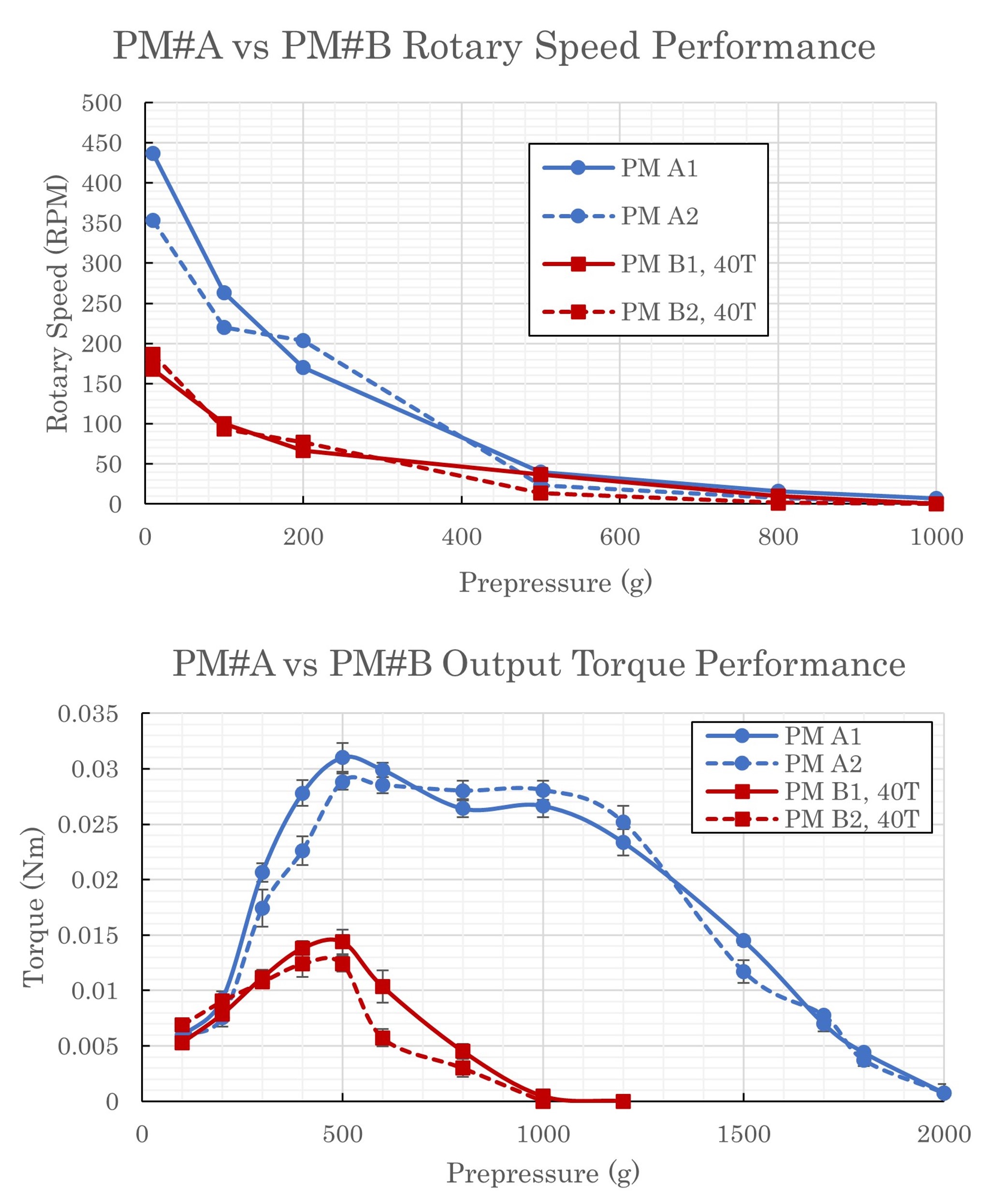}
    \caption{Speed and torque performance of group A and B motor compared.}
    \label{fig:PMAB}
\end{figure}

Group B motors reduced the teeth number to 40 compared to group A motors with 48 teeth. A detailed parameters description can be found in Figure \ref{fig:PMB}. Figure \ref{fig:PMAB} shows the results of the speed and torque of groups A and B compared. Results indicated that group A with 48 teeth was better than group B with 40 teeth in torque. Group B motors performed 180RPM speed and 0.0151Nm torque, which were both lower than group A motors' performance. Similar to group A motors, the highest speed was performed at initial prepressure at 10g, and the highest torque was performed at approximately 500g, however, both motors stopped spinning completely when the prepressure reached over 1000g. 

\subsubsection{Group C: Chamfer on the Teeth}

\begin{figure}
    \centering
    \includegraphics[width=0.8\linewidth]{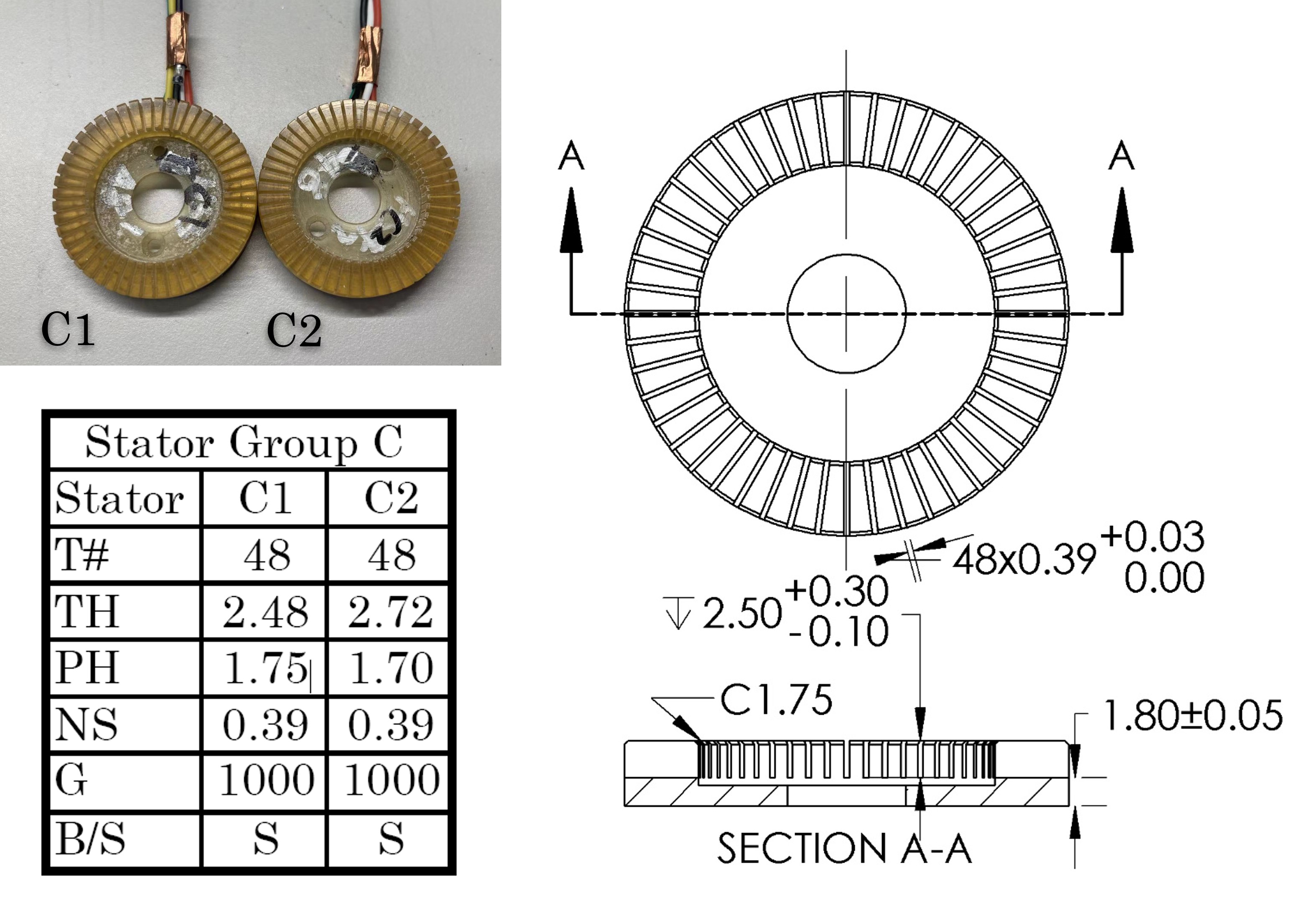}
    \caption{Group C plastic stators condition, including image, dimensional drawing, and parameters.}
    \label{fig:PMC}
\end{figure}

\begin{figure}
    \centering
    \includegraphics[width=0.8\linewidth]{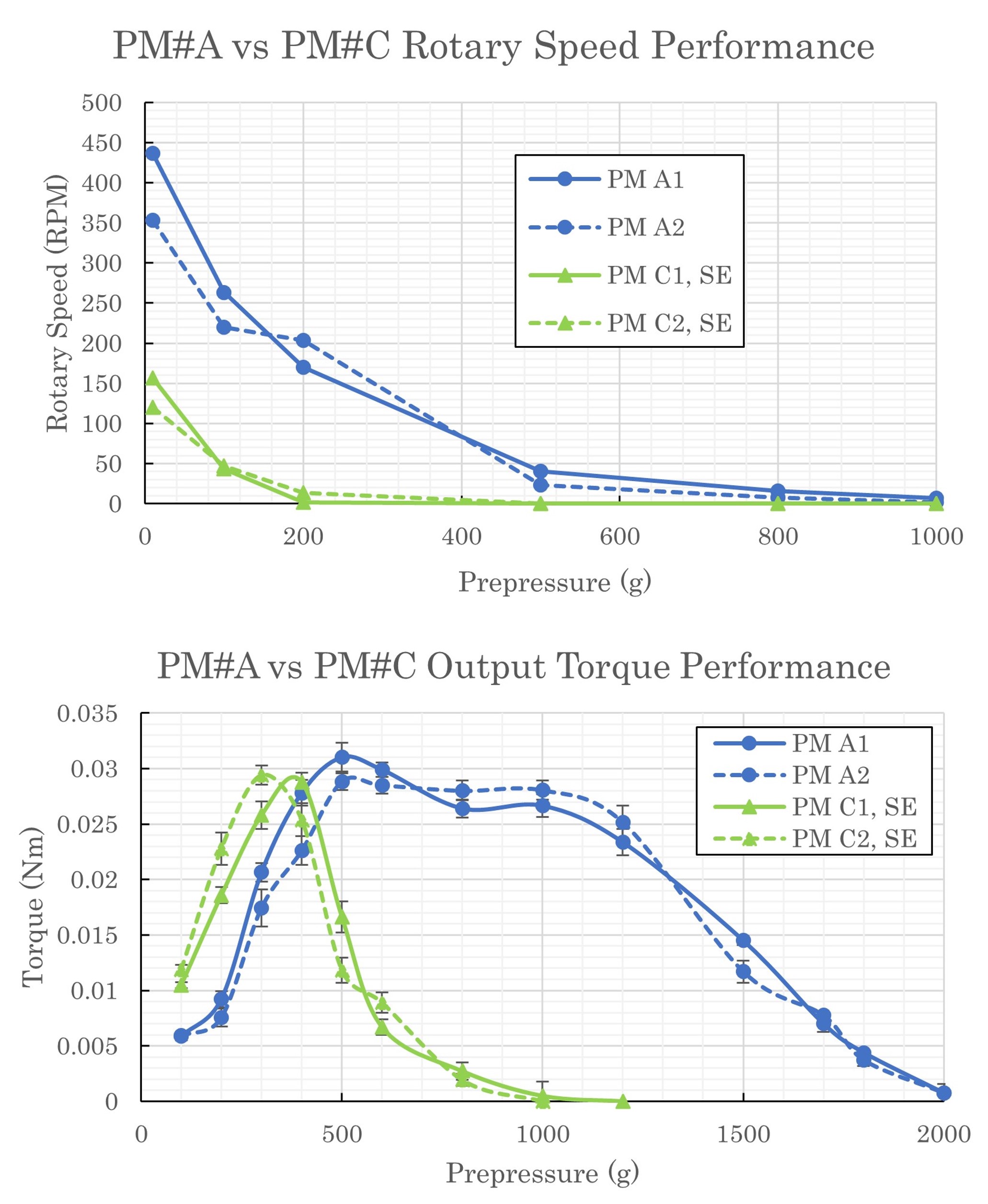}
    \caption{Speed and torque performance of group A and C motor compared.}
    \label{fig:PMAC}
\end{figure}

Group C motors remained the straight edge without a beveled edge compared to group A motors. A detailed parameters description can be found in Figure \ref{fig:PMC}. Figure \ref{fig:PMAC} shows the results of the speed and torque of group A and C motors compared. Results show that group A with a beveled edge was better than group C with a straight edge in speed aspect, where 150rpm was measured, however, group B motor performed larger torque at the initial prepressure, where the largest prepressure was shifted to 300g, and 400g and performed 0.030Nm and 0.025Nm torque respectively. Then the torque dropped significantly when it reached 500g and above, and similar to group B motors both motors stopped spinning completely when prepressure reached 1000g and above.

\subsubsection{Group D: Teeth Width}

\begin{figure}
    \centering
    \includegraphics[width=0.8\linewidth]{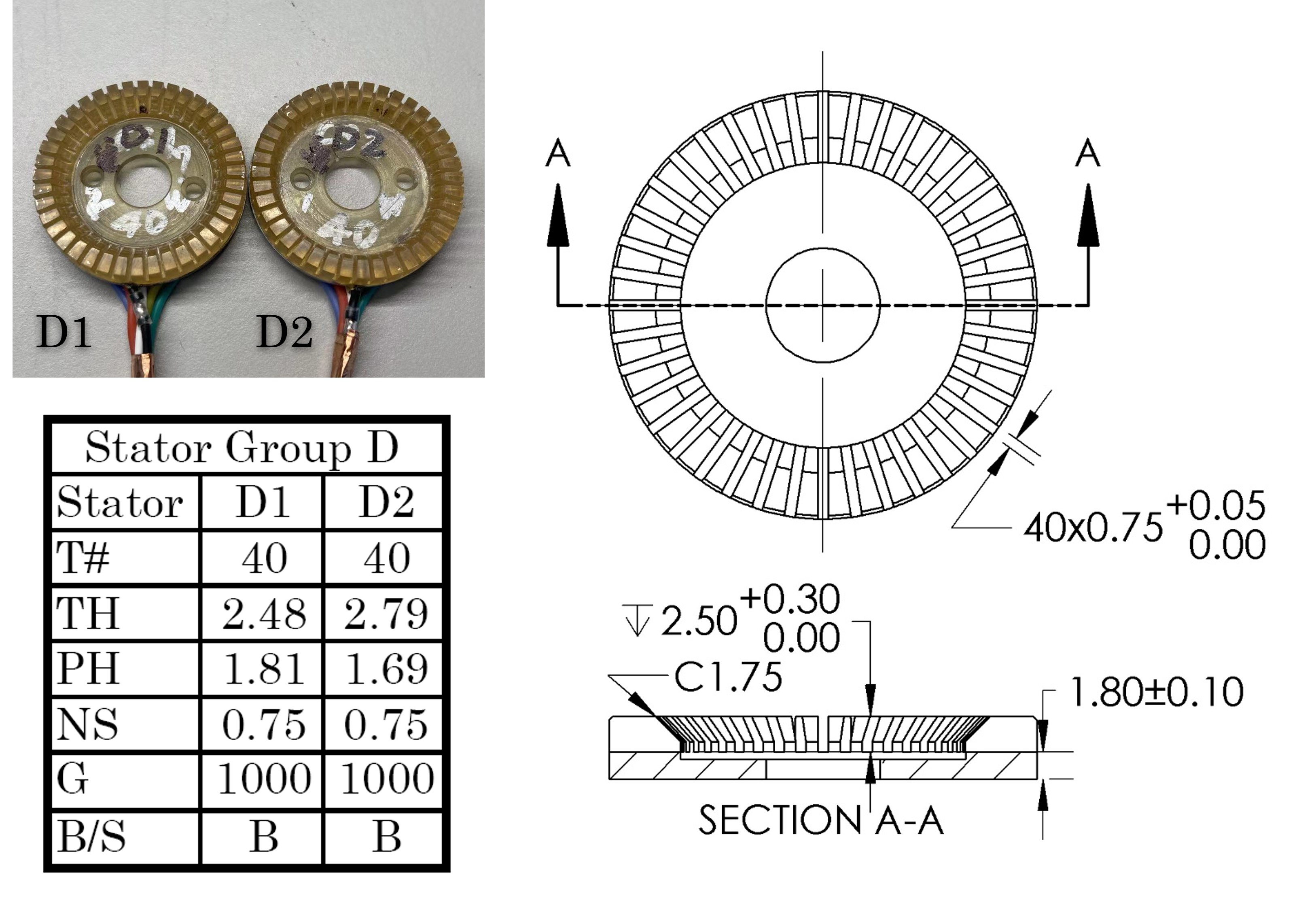}
    \caption{Group D plastic stators condition, including image, dimensional drawing, and parameters.}
    \label{fig:PMD}
\end{figure}

\begin{figure}
    \centering
    \includegraphics[width=0.8\linewidth]{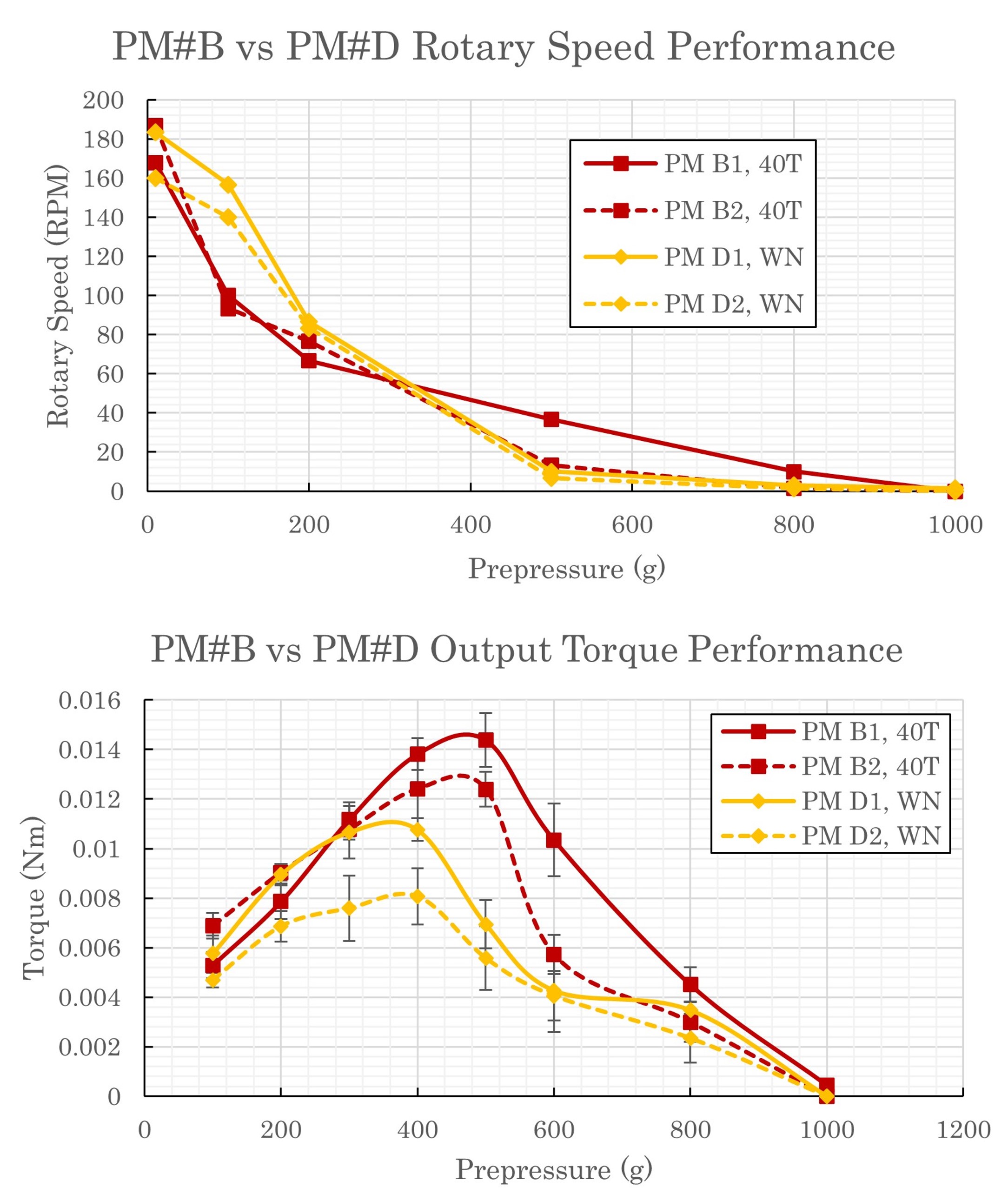}
    \caption{Speed and torque performance of group B and D motor comparing.}
    \label{fig:PMBD}
\end{figure}

Group D motors changed the notch width to 0.75mm compared to other groups with 0.39mm, which indicates a shrinkage of teeth width, and the description of a detailed parameter can be found in Figure \ref{fig:PMD}. Considering the shrinkage of teeth width will cause an extremely small contacting area between the stator and rotor, also thin teeth will cause melting while manufacturing the stators, so we chose 40 teeth for group D and compared them with group B motors. Figure \ref{fig:PMBD} shows the results of the speed and torque of groups B and D compared. Results indicated that groups B and D acted with similar performance both in speed and torque. Specifically, group B performed higher speed at 100g prepressure, approximately 160rpm from group D versus 100rpm from group B. However the torque performance of D was even worse compared to B, where the highest torque namely 0.011Nm with prepressure value shifted to 300-400g.

\subsubsection{Group E: Surface Roughness}

\begin{figure}
    \centering
    \includegraphics[width=0.8\linewidth]{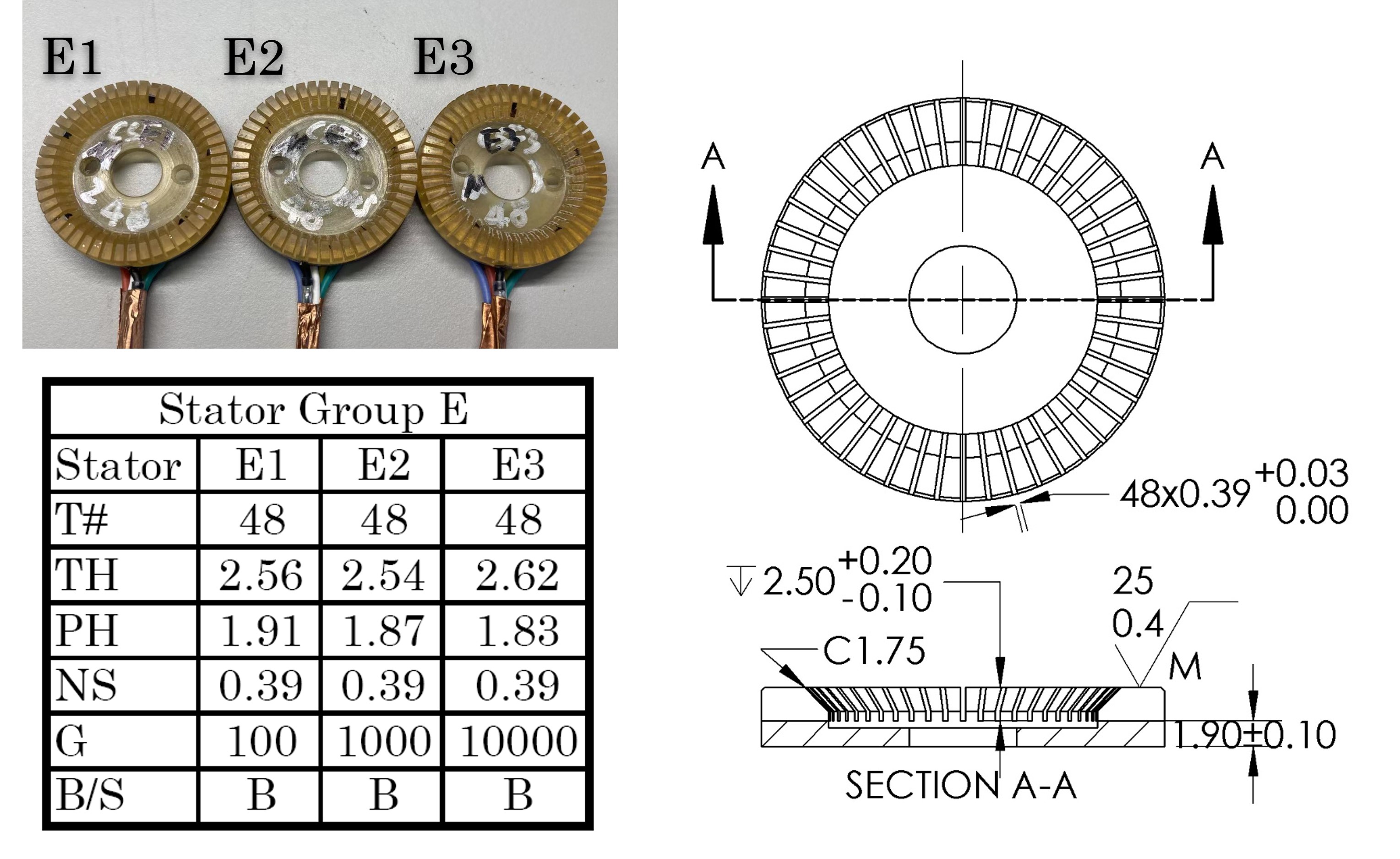}
    \caption{Group E plastic stators condition, including image, dimensional drawing, and parameters.}
    \label{fig:PME}
\end{figure}

\begin{figure}
    \centering
    \includegraphics[width=0.8\linewidth]{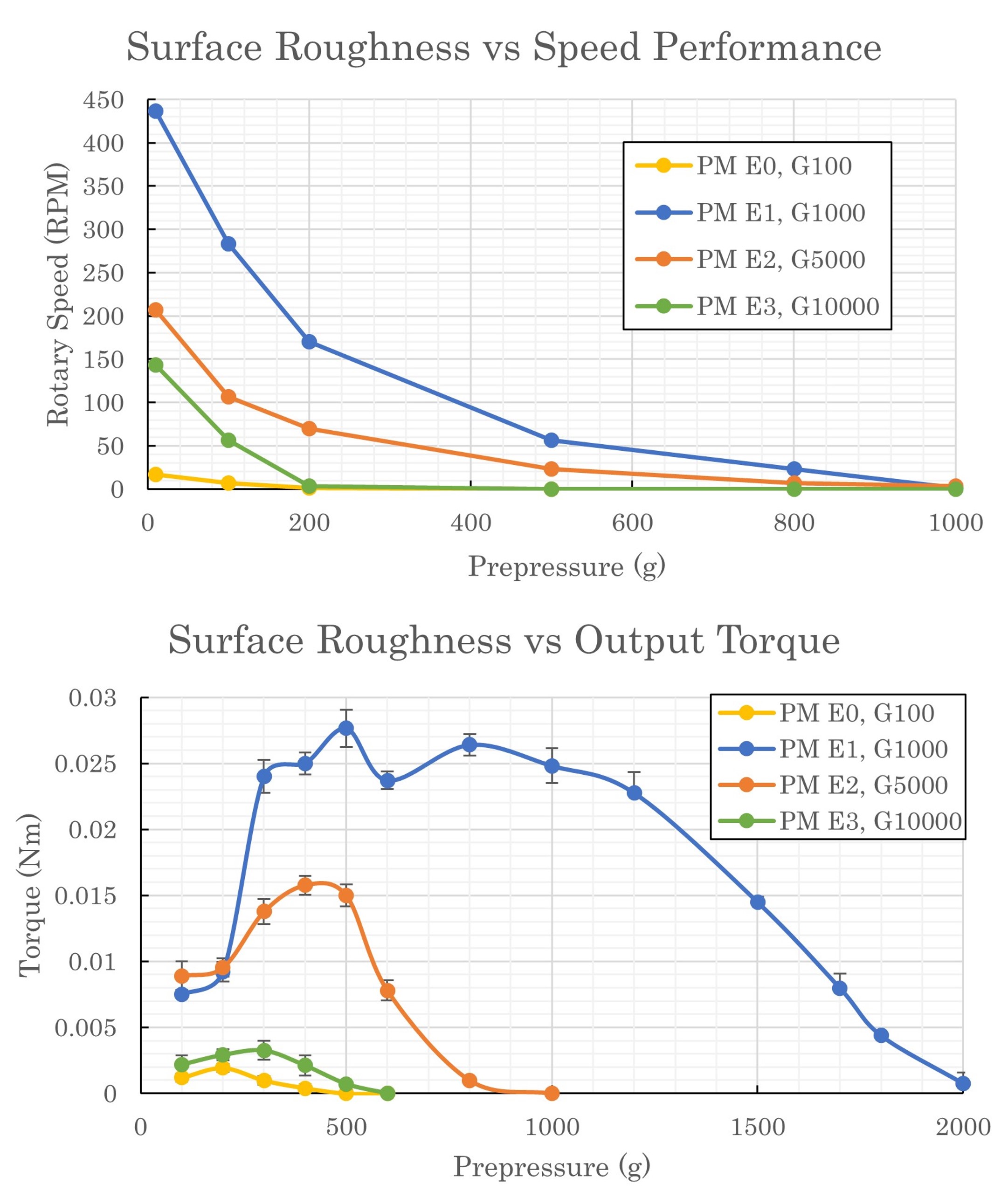}
    \caption{Speed and torque performance of group E.}
    \label{fig:PMEresults}
\end{figure}

Group E motors consisted of 4 different levels of surface roughness and were compared with each other within the group. A detailed parameters description can be found in Figure \ref{fig:PME}. We used grit 100, 1000, 5000, and 10000 sandpapers to create different levels of surface roughness finishing. Note that the E1 stator is equivalent to group A stators. Figure \ref{fig:PMEresults} shows the results of the speed and torque performance of group E. Results indicate that the surface finish level polished by grit 1000 sandpaper will generate both the best performance of speed and torque, where the highest values are 430.7666rpm and 0.029Nm respectively.

\subsubsection{Motor Design Suggestion}

\begin{figure}
    \centering
    \includegraphics[width=0.9\linewidth]{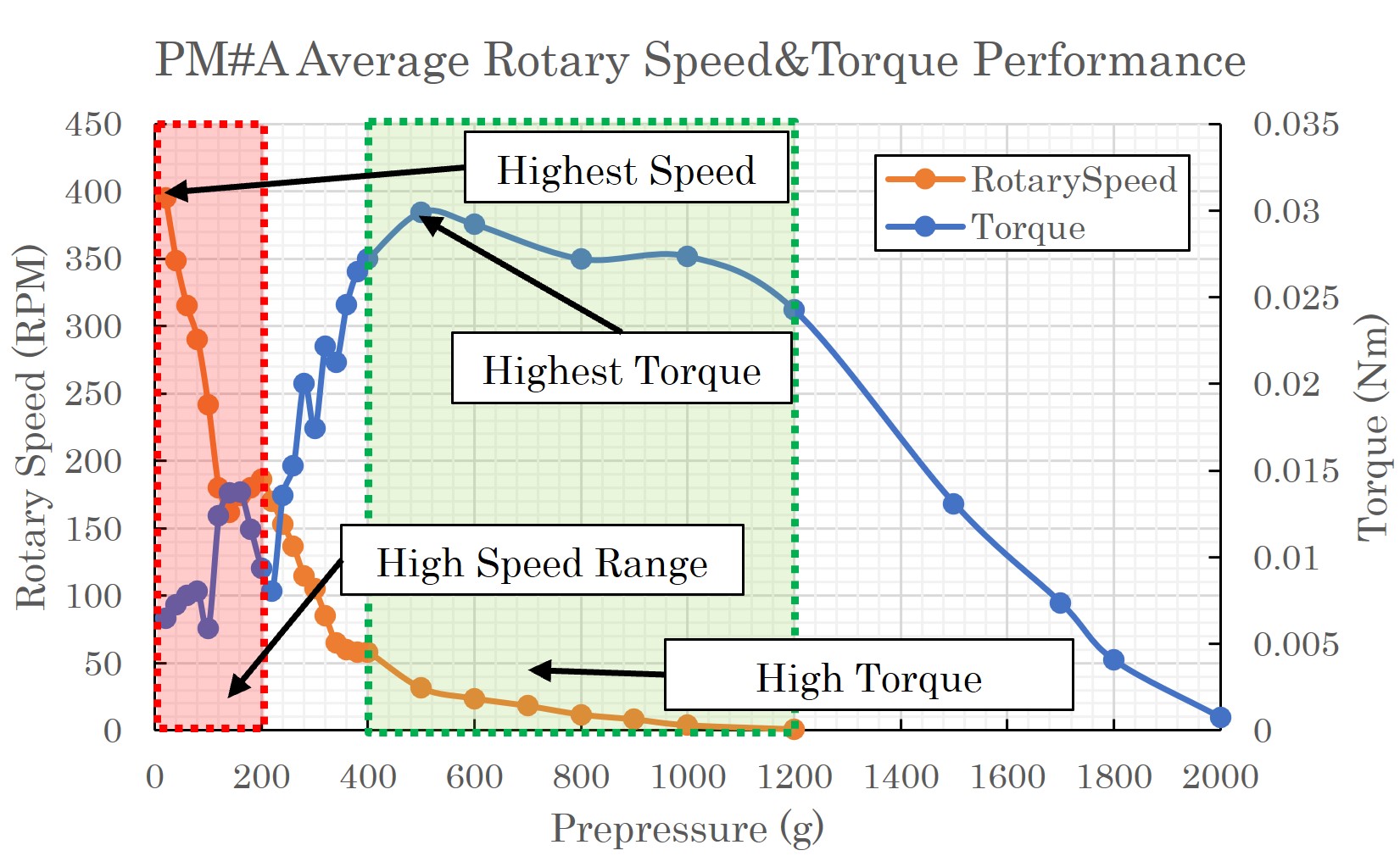}
    \caption{Optimized motor design outputting performance versus prepressure.}
    \label{fig:PMdata}
\end{figure}

Based on all the experiments and measured data above, we can get the optimized motor design outputting performance versus prepressure in gram, which is shown in Figure \ref{fig:PMdata}. Because we proved that the group A motors performed the best value compared to other groups, so we used an average of A1 and A2 motor data in this image. For the Ultem plastic motor, pressure is much smaller compared to the copper-made stator, the high-speed range performs from 10 to 200g prepressure, and the high torque range performs from 400 to 1200g. Specifically, the highest speed is up to 436.6665rpm when the prepressure is low, and the highest torque is up to 0.0348Nm when the prepressure is approximately 500g. Also, the best output performs at a specific balance point of surface roughness, namely the finishing using grit 1000 sandpaper.

\section{Conclusion and Disscussion}

In this study, we presented a fully plastic motor design and evaluated its performance within an MRI environment, focusing on the mean SNR drop. Our experimental results allowed us to identify an optimized parameter combination for the stator in the current stage of the plastic motor design. The findings indicated no significant change or statistical difference in the plastic motor's performance when stationary, compared to baseline conditions. However, when the motor was spinning, T1 and T2-weighted scans revealed a statistical SNR reduction of approximately 13\% and 16\%, respectively. These SNR reduction values represent an improvement over the plastic-enclosed motor reported in \cite{carvalho2020demonstration}.

We explored various parameter variations and validated the rotary speed and output torque through experimentation, ultimately identifying an optimal design parameter combination. The parameters tested included teeth number, notch size, edge design (beveled or straight), and surface finish quality versus prepressure. Results indicated that a configuration with 48 teeth, a thin notch of 0.39mm, beveled edges, and a surface finish achieved with approximately 1000-grit sandpaper provided superior performance in terms of rotary speed and torque. Under this configuration, the motor reached a maximum speed of 436.67 rpm at low prepressure and a peak torque of 0.0348 Nm at a prepressure of approximately 500g.

The current design and parameter combination suggest a functional plastic ultrasonic motor (USM). However, material selection was not covered in this work and remains an area for future study. For instance, materials like Macor or quartz could be considered for stator fabrication to enhance performance. Additionally, the role of the friction layer, particularly in USMs, warrants further investigation. While metal stator-rotor systems typically require a Teflon layer to increase friction, replacing copper with Ultem 1000 plastic results in a softer surface, potentially altering the need for such a layer. Further studies could explore whether an additional Teflon layer is necessary, a topic related to surface roughness.

Future work could also focus on optimizing motor geometry, including teeth configuration and plate height. Theoretical studies employing sequential quadratic programming or topology-based sensitivity analysis might offer insights into the optimal geometry. For instance, the ratio of teeth to notch could be simulated across different ranges to evaluate performance. Motor geometry optimization guidelines can be found in \cite{zhang2012optimal}. Additional research could also aim at further refining the motor's overall geometry to advance towards a semi-commercial MRI-compatible rotary motor, integrate feedback functions into the current custom motor design, and prepare for the next generation of MRI-compatible robot design and experimentation.

\bibliographystyle{IEEEtran}
\bibliography{Main}

\end{document}